\documentclass[sigconf]{acmart}

\AtBeginDocument{%
  }

\setcopyright{rightsretained}
\copyrightyear{2024}
\acmYear{2024}

\acmDOI{}

\acmConference[KDD CJ Workshop '24]{Workshop on End-to-End Customer Journey Optimization}{August 25,
  2024}{Barcelona, Spain}
\acmISBN{}

\begin{document}

\title{Adaptive User Journeys in Pharma E-Commerce with Reinforcement Learning: Insights from SwipeRx}

\author{Ana Fernández del Río}
\email{ana@causalfoundry.ai}
\affiliation{%
  \institution{Causal Foundry}
  \city{Barcelona}
  \country{Spain}
}

\author{Michael Brennan Leong}
\email{leong@swiperxapp.com}
\affiliation{%
  \institution{SwipeRx}
  \city{Jakarta}
  \country{Indonesia}
}

\author{Paulo Saraiva}
\email{paulo@causalfoundry.ai}
\affiliation{%
  \institution{Causal Foundry}
  \city{Barcelona}
  \country{Spain}
}

\author{Ivan Nazarov}
\email{ivan@causalfoundry.ai}
\affiliation{%
  \institution{Causal Foundry}
  \city{Barcelona}
  \country{Spain}
}

\author{Aditya Rastogi}
\email{aditya@causalfoundry.ai}
\affiliation{%
  \institution{Causal Foundry}
  \city{Barcelona}
  \country{Spain}
}

\author{Moiz Hassan}
\email{moiz@causalfoundry.ai}
\affiliation{%
  \institution{Causal Foundry}
  \city{Barcelona}
  \country{Spain}
}

\author{Dexian Tang}
\email{dexian@causalfoundry.ai}
\affiliation{%
  \institution{Causal Foundry}
  \city{Barcelona}
  \country{Spain}
}

\author{África Periáñez}
\email{africa@causalfoundry.ai}
\affiliation{%
  \institution{Causal Foundry}
  \city{Barcelona}
  \country{Spain}
}

\renewcommand{\shortauthors}{Fernández del Río and Leong, et al.}

\begin{abstract}

This paper introduces a reinforcement learning (RL) platform that enhances end-to-end user journeys in healthcare digital tools through personalization. We explore a case study with SwipeRx, the most popular all-in-one app for pharmacists in Southeast Asia, demonstrating how the platform can be used to personalize and adapt user experiences. Our RL framework is tested through a series of experiments with product recommendations tailored to each pharmacy based on real-time information on their purchasing history and in-app engagement, showing a significant increase in basket size. By integrating adaptive interventions into existing mobile health solutions and enriching user journeys, our platform offers a scalable solution to improve pharmaceutical supply chain management, health worker capacity building, and clinical decision and patient care, ultimately contributing to better healthcare outcomes.

\end{abstract}




\maketitle

\section{Introduction}

Personalization and adaptability are crucial for enhancing user experience and customer journeys in digital tools. Leveraging detailed user in-app behavior data, reinforcement learning (RL) provides adaptive journeys that improve user experience. This is particularly relevant for mobile health solutions, especially for healthcare workers in low- and middle-income countries (LMICs), where such tools can help mitigate the lack of resources. Pharmacies play a critical role in healthcare systems everywhere. They are often the only primary healthcare contact in LMICs for many patients while facing barriers like a shortage of trained professionals, medication, and suboptimal supply chains~\cite{Miller2016, Ikhile2018, Hamid2020}. Supporting pharmacists with inventory management, easy-to-access clinical guides, and connection to their peers and the rest of the health system can significantly impact their communities.

\subsection{The Reinforcement Learning Platform}
\label{sec:platform}

\begin{figure*}
    \centering
    \includegraphics[width=0.9\linewidth]{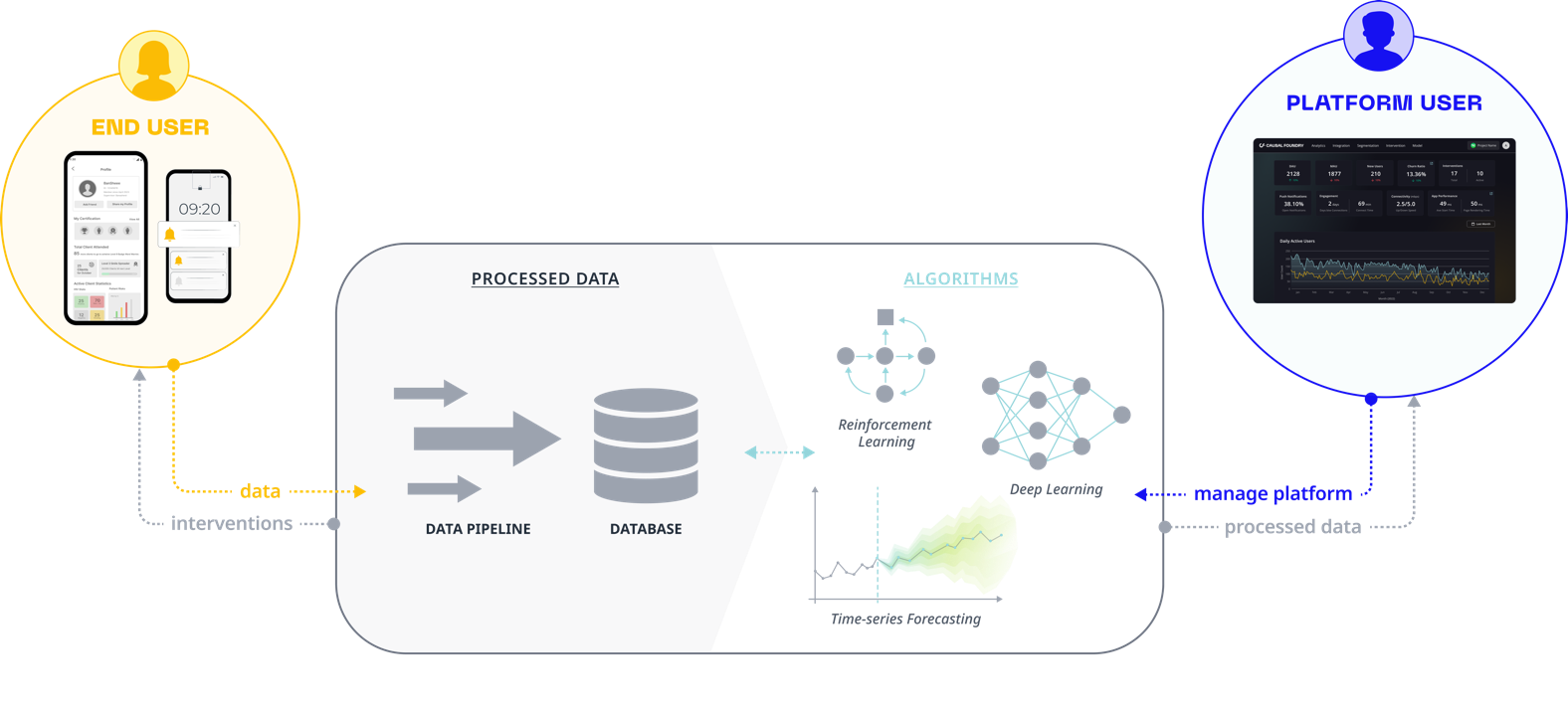}
    \caption{Schematic representation of the RL platform's architecture.}
    
     \Description{Data from the app's end users is tracked by the SDK and processed by the platform's data pipelines to be used by the predictive and intervention algorithms, which can be managed and visualized by the RL platform's user through its frontend. Interventions are delivered to app users also through the SDK}
    \label{fig:platform}
\end{figure*}

We propose an artificial intelligence (AI) data-centric platform that can integrate into already existing healthcare-related digital tools to enhance them with an adaptive user journey through reinforcement learning (RL). By enabling the system to learn and adapt its behavior based on the feedback it receives, RL is used to optimize the user experience, personalizing it and making it more efficient. 
The end-to-end machine learning (ML) platform can integrate into different healthcare related software and leverages the behavioral and clinical logs from these solutions, together with other contextual information sources, to deliver adaptive interventions 
directly to their recipients through these tools in the form of personalized recommendations, reminders, incentives, or in-app content and workflows, providing an adaptive user experience and journey. 
We will illustrate the framework and technologies proposed by focusing on a pharmacist-facing tool, particularly its bussiness-to-bussines (B2B) e-commerce component.

Figure \ref{fig:platform} depicts the platform's architecture, which consists of a \emph{Software Developer Kit} (SDK) which is embedded into the digital tools to track labeled data and deliver interventions, a \emph{frontend} interface with analytics, model management, intervention and experimentation functionalities, and the \emph{backend}, which takes care of log ingestion, data transformation, scheduled dispatch of nudges (message or content interventions) and hosts the algorithmic machine learning (ML) engine. The platform has also been described in~\cite{perianez2024} and has been designed to integrate with mobile health solutions beyond the pharma supply chain, including patient management~\cite{epidamik} and clinical decision support, facility logistics, patient support, and patient management. A shorter version of this paper is also available~\cite{behavioral}.

\subsubsection{SDK}

One of the platform's core functionalities is data tracking, organizing, and labeling. Ensuring data collection and its quality is arguably the most critical step in moving towards a solid AI approach to personalization in digital health. The standardized design of event logging enables the platform to be a uniform interface for transformation and aggregation across different application domains, e.g., supply chain and e-commerce marketplace, medication tracking, and patient communication. The SDK is a well-defined library and data structure at the heart of the platform's data side. It provides the tools for collecting relevant events on an edge device (e.g., mobile, wearable monitor, or smart device) and the messaging service that delivers the interventions. 

\subsubsection{Backend}
\label{sec:backend}

The incoming logs through the SDK are processed by the tailor-made data pipeline and categorized into dynamic and static traits, which can aggregate through arbitrary time resolution. These traits represent insights about the users and their interactions with different in-app content and other subjects (pharmacies, patients, drugs), and once derived, they become available throughout the platform. They can be used to track behavior and outcomes, to group subjects or content (e.g., drugs or CPD modules) into meaningful cohorts (e.g., for pharmacy segmentation), as features/covariates for statistical and predictive modeling, or to make up the contexts and rewards for bandit-based decision algorithms. The backend orchestrates the storage and processing of domain-knowledge analytical traits, the computation of features and predictions derived from traits using advanced machine learning, and, finally, issues personalized nudges back to SDK instances.

\subsubsection{Algorithmic engine}
\label{sec:algorithmic-engine}

The machine learning component hosted in the backend is composed of analytic and predictive modeling, an ML recommendation engine, and an algorithmic decision-making service. These modules configure, train, host, and manage models for statistical analysis, time series forecasting, deep and ensemble survival analysis, item recommendations, and sequential decisions via reinforcement learning (see Section \ref{sec:methodology} for more technical details). These services expose their functionality via JSON request-response API to the backend. 

\subsubsection{Frontend}

The frontend provides an intuitive user interface for configuring models and interventions and an analytics dashboard for monitoring ongoing ones. The configuration UI guides the user from the subject cohort and sample definition through algorithm selection to feature selection and target specification, including nudge alternatives. The dashboard represents various traits, predictions, and metrics in a convenient format for easier interpretation of the results.

\subsection{SwipeRx}

\begin{figure}
    \centering
    \includegraphics[width=0.4\linewidth]{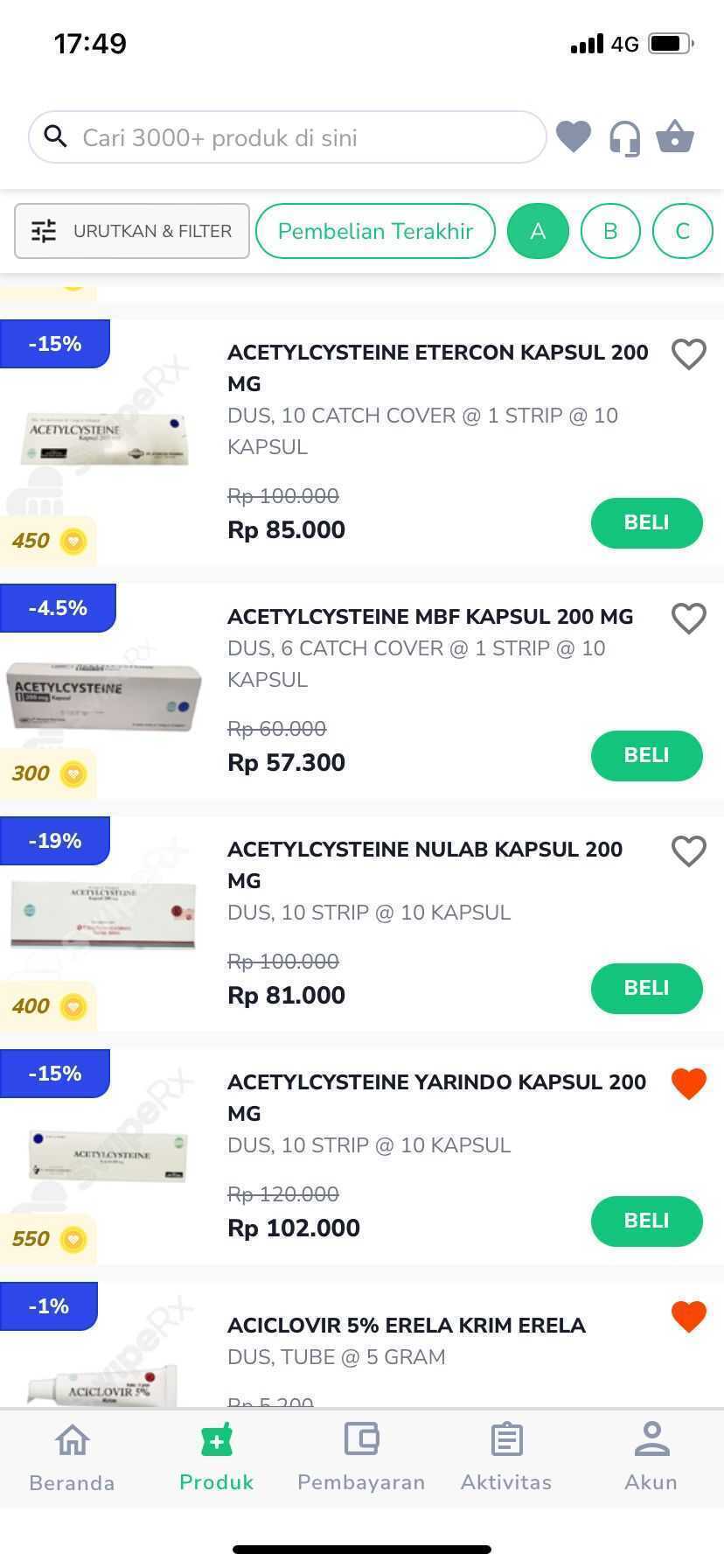}
    \includegraphics[width=0.4\linewidth]{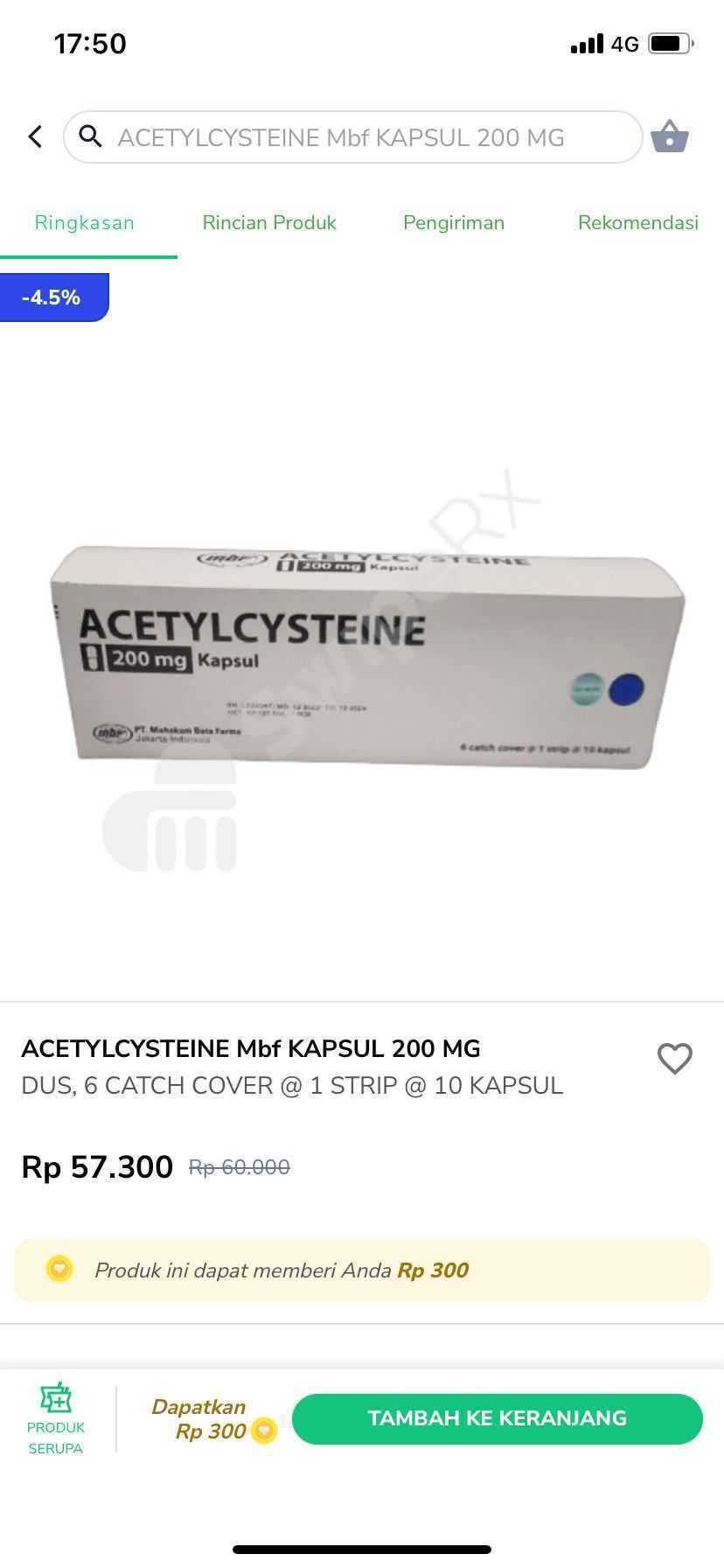}
    
    \Description{Screenshots of the SwipeRx application e-commerce section}
    \caption{Screenshots of the SwipeRx application.  }    \label{fig:screenshots}
\end{figure}

SwipeRx, the largest network of pharmacies in Southeast Asia, connects over 235,000 professionals across more than 45,000 pharmacies, revolutionizing the industry with its community-driven commerce model in Indonesia, the Philippines, Malaysia, Thailand, Cambodia, and Vietnam. The platform provides comprehensive services, including online education, centralized purchasing, logistics and financing, news, drug directory, adverse event reporting, and more, addressing the unique needs of this vital public health sector. Figure \ref{fig:screenshots} shows some screenshots of the app.

The network is especially crucial in the region, where there's a heavy reliance on pharmacies due to a scarcity of professional physicians (with patients visiting pharmacies ten times more than doctors every year). 
Through its app, SwipeRx fosters knowledge sharing and collaboration among pharmacists, enhancing the quality of patient care in local communities. Over 80,000 pharmacy professionals have already been educated through its digital professional education.

In Southeast Asia, more than 80\%  of the pharmacies are independent, facing challenges of quality, availability, and affordability of essential medicines. SwipeRx is helping to solve this by empowering pharmacy professionals with the technologies and tools they need and addressing the operational challenges these predominantly small, family-owned pharmacies face, such as financing and supply chain management. By facilitating bulk purchases and analyzing drug consumption data, SwipeRx enables these pharmacies to procure medicines at competitive prices, ensuring they can meet patient needs effectively. 

\subsection{Adaptive User Journey}

We propose a user-centric approach where each user's journey is tailored to their evolving needs, preferences, and the exact stage of their relationship with the service or tool to ensure that every user's unique requirements are met. This is made possible by the RL platform's adaptive algorithms (see Section \ref{sec:bandits}) ability to systematically use the information from all users and continually adapt to improve user experience.

It begins with a personalized onboarding process by adjusting the level of support and introduction to app functionalities according to each user’s requirements. An adaptive algorithm can determine the complexity of tips offered or whether to provide tips at all, based on user data such as location, experience, gender, job title, app usage patterns, and previous onboarding interactions. This ensures that tech-savvy users can navigate the app unhurdled by unnecessary popups, while less experienced users receive step-by-step guidance. An adaptive onboarding experience can also adjust workflow complexity based on the user's expertise. For example, community health workers (CHWs) struggling with technology are given a simplified patient screening workflow to keep them engaged without overwhelming them while collecting essential information. As they gain experience, the workflow introduces more complex options, allowing experienced users to provide detailed patient data, referrals, clinical history, or preventive education provided.

Beyond onboarding, the personalized journey continues with tailored in-app content and workflows and with messages (via popups, notifications, WhatsApp or SMS) with personalized motivational prompts, tips, discounts, and incentives, to foster long-term user engagement. In tools such as SwipeRx, messages with personalized product recommendations (like those discussed in Section \ref{sec:adaptive-item-pair-recommendations-pharmacies}) and in-app ML-ordered product lists help increase or maintain purchasing engagement. The adaptive delivery mechanisms, in this case, can also use the outputs of predictive models to be more efficient, such restocking recommendations based on demand prediction that can also help reduce stockouts of essential drugs. Churn prediction models also play a significant role, as they can be used to precision target users at high risk with specific interventions, such as special discounts. Specific adaptive algorithms (see Section \ref{sub:restless_bandits}) able to allocate limited resources such as discounts or even a call from a sales representative efficiently.  

\section{Methodology}
\label{sec:methodology}

The following subsections describe in some technical detailthe platform's algorithmic server's (Section \ref{sec:algorithmic-engine}) capabilities.

\subsection{Predictive Modeling}

Predictive modeling can help to define who should be targeted with interventions and the right way and moment to do so.  Note that while predictions are not used in the first set of interventions described in Section \ref{sec:adaptive-item-pair-recommendations-pharmacies}, we include this subsection for completeness as different predictions are readily available in the platform.

The platform includes different families of predictive models suitable for different use cases. Through its frontend, platform users can easily create models from these families by designating the traits to be used as targets and features and the specific algorithm with its parameter values. The model lifecycle can also be managed through the platform, including training schedules and performance monitoring. All resulting outputs of the models, including estimations of uncertainty and feature importance, are available to create additional traits that can be used to characterize subjects and content throughout the rest of the platform.

Survival analysis approaches~\cite{Wright2017, Fu2016, Lee2020, olaniyi2022user} offer a way of characterizing user (pharmacy, patient, distribution center) behavior by modeling the evolution with time of the probability of occurrence of different events of interest. Formally, the methods learn a function $g \colon \mathcal{X} \times [0, +\infty) \to [0, 1]$ from the observed data $(x_j, t_j, c_j)_{j=1}^m$ with the input features $x_j \in \mathcal{X}$, censored time-to-event $t_j \in (0, M]$, and the censoring indicator $c_j \in \{0, 1\}$, so that $g(x, \cdot)$ approximates the conditional survival function $P(T \geq \cdot \mid X = x)$ of the duration until the event of interest conditional of the input features $x \in \mathcal{X}$.
They can be used, for example, to predict the risk of non-adherence to treatment, of pharmacy churn in an e-commerce platform, or of a pharmacist failing to complete a CPD module, which would be indicating adequate subjects to target with specific interventions (e.g., reminders to support treatment adherence, promotions to discourage churn, motivational prompts and incentives to encourage learning).

Demand predictions are crucial in e-commerce, supply chain, and inventory optimization. Multivariate forecasting methods that learn either point or probabilistic forecasts from co-dependent time series are viable solutions for predicting the demand for different products at different sites~\cite{benidis_deep_2022}. These approaches include autoregression- and graph-regularized factor analysis~\cite{yu_temporal_2016} and latent state forecasting~\cite{seeger_approximate_2017}, state space modeling with LSTM and Gaussian copula for probabilistic forecasts~\cite{salinas_high-dimensional_2019}, recurrent networks for multivariate count data with non-uniform scale~\cite{salinas_deepar_2020}, and attention-based deep models for multi-horizon interval predictions~\cite{lim_temporal_2021}.

\subsection{Recommendation Algorithms}

As will be described in section \ref{sec:bandits}, the main mechanism in place to learn and exploit user preferences for recommendation and personalization will be that of adaptive delivery, i.e., bandit-based to a large extent. Other recommendation algorithms, however, might provide valuable insights that can be leveraged by the reinforcement learning algorithms and used for content recommendations. 
The RL platform includes a neural embedding based item-to-item recommender~\cite{grbovic2015, barkan2016item2vec, eide2018}, in which items are recommended based on their similarity to other items.  This similarity is based both on the fixed known information about each item and on the interaction history of users with them. Transformers are used to convert available features written in natural language (e.g., name, description, manufacturer, ingredients...) into numerical vectors. Principal component analysis (PCA) is then applied to reduce their dimensionality. The PCA-transformed natural language vectors together with the interaction history (e.g., purchase orders item content) are fed into a fully connected neural network with three layers. The output of this system are the L2-normalized embedding characterizing each of the items, whose similarity to other items can be then computed as the Euclidean distance between their embeddings.
However, the set of interventions discussed in this paper relies on the simpler rule-based algorithm described below.

\subsubsection{Item-pair recommendation}
\label{sec:item-pair-recommendation}

The underlying idea is to find products that the pharmacy is very likely already purchasing, albeit elsewhere, and make them aware that these are also part of SwipeRx's catalog. This is achieved by recommending pairs of products which are typically purchased together by the user population to pharmacies that are only ordering one of the items frequently.

Formally, consider the pair $p_k = (j,j)$, where $i$ and $j$ represent products. Define $d_{uit}$ as at time $t$, the number of days since user $u$ purchased product $i$. Let $\bar{d}_{uitT}$ be at time $t$ the average number of days between purchases in the last $T$ months. For the initial intervention with SwipeRx, $T$ was chosen to be three months. If $i$ has never been purchased by $u$, we adopt the convention that $d_{uit}=-1$. The recommendation follows a two-step process. In the first step, the list of candidate pairs is generated. Namely, the top 100 pairs purchased together are selected by ranking them according to either the number of times they have been purchased together or the generated revenue in the past $T$ months. For the initial intervention with SwipeRx, revenue was chosen. Keep in the list only pairs in which both items are in stock. In the second step, user-specific filtering is made. First, for user $u$, retain the list of $p_k=(i,j)$'s such that, without loss of generality, $i$ has been purchased recently, defined as $d_{uit}/\bar{d}_{uitT} \in (0,1)$. For user $u$, pick $j$ such that $d_{ujt} = -1$ or $d_{uit}/\bar{d}_{uitT} \notin (0,1)$. Retain the pairs such that $d_{ujt}=-1$. If no such $j$ exists for user $u$, retain pairs with the highest values for $|d_{uit}/\bar{d}_{uitT}-d_{uit}/\bar{d}_{uitT}|$, i.e., a large difference in expected recency between $i$ and $j$. Finally, from the ultimately finalized list of pairs, one is selected at random.

\subsection{Bandits and Adaptive Intervention Delivery}
\label{sec:bandits}

The proposed framework aims to leverage data to intervene and nudge behavior. As such, the algorithmic piece deciding what interventions to use (or not), and when for each user is central. The intervention decision problem can be modeled as a Markov Decision Process (MDP), and reinforcement learning is the appropriate paradigm to drive personalization. It allows balancing between optimization and knowledge extraction depending on the use case and continually adapts to the evolving needs and preferences of the intervention subjects. 

Stochastic contextual and restless multiarmed bandits (MABs) are good frameworks for instances of personalization and resource allocation where the long-term state evolution plays a minor role, such as those applications where we are concerned mainly with its immediate impact, or where simple, useful approximations to state dynamics can be found.
Problems that need to robustly reconcile short and long-term goals through sequential decisions that optimize for the multi-step problem, such as decisions associated with medical treatment of certain conditions, require frameworks that model the full MDP~\cite{Liu2020, Yu2019}, in the spirit of collaborative, interactive recommendation systems~\cite{Ie2019, Liu2019, Lillicrap2019, Chen2020}, or at least address the problems of long-term credit assignment, when learning the intervention policy.

Bandit algorithms can be thought of as \emph{online optimization methods} with \emph{built-in model identification} from \emph{partial feedback}: the algorithm must make sequential decisions based on the observed interaction history with the ultimate goal of eliminating suboptimal choices~\cite{lattimore_bandit_2020}. During the course of $T$ repeated interactions, the algorithm observes a context $x_t$, then, given the history of its past contexts, interactions, and their outcomes $\mathcal{H}_{<t} = \{x_s, a_s, r_s\}_{s < t}$, it picks an action $a_t$ from a finite set. It concludes the interaction by receiving a response-reward $r_t$. In this context, the action $a_t$ represents the intervention decision at time $t$, and both the context $x_t$ and reward $r_t$ are chosen from within the static and dynamic traits described in section \ref{sec:backend}.
The (contextless) MABs are a special case, wherein the context is empty, i.e., an interaction-independent constant~\cite{lattimore_bandit_2020}.

During its rollout, a bandit algorithm gradually hones in on a set of optimal actions with a guaranteed confidence level for the outcome, allowing it to eventually abandon sub-optimal actions. This allows it to eventually minimize the gap between the taken and the unknown optimal action (the \emph{regret}).

It is assumed that the law $r_t \sim p(r \mid a_t, x_t)$ governing the interaction outcome is unknown but fixed and that current actions have no effect on the future contexts and interactions~\cite{lattimore_bandit_2020}. This is in contrast to the full MDP reinforcement learning setting, where the object being interacted with is \emph{a persistent state, which evolves due to interactions under a Markov decision process}~\cite{sutton_reinforcement_2018}.

An important distinction from supervised or active learning is that bandit (and reinforcement learning) methods observe only partial feedback, meaning that the opportunity costs when learning an optimal policy are taken into account. As such, these algorithms need to strike a balance between \emph{exploration}, that is, reliable identification of the action-reward feedback, and \emph{exploitation}, i.e., leveraging the learned model to action policy that achieves the objective optimization goals.

There are extensions of the bandit setting to very large action spaces encountered in recommender systems~\cite{lopez_learning_2021}, to settings in which the optimization aspect is not as crucial as the best-arm identification capabilities within a strict budget~\cite{bubeck_pure_2011}, and situations with budgeted interactions with a batch of concurrent Markov processes, which will be described in Section~\ref{sub:restless_bandits}.

The following subsections cover the different algorithms available in the RL-platform described in \ref{sec:platform} for operational sequential decision making.

\subsubsection{Linear Bandits}
\label{sub:linear_bandits}

A $k$-armed linear bandit assumes a linear model of the reward conditional on a context-action pair: $r_t = x_t^\top \theta_{a_t} + \varepsilon_t$, where $x_t \in \mathbb{R}^d$ is the feature vector, $(\theta_k)_{k=1}^K \in \mathbb{R}^d$ are the coefficients of the $k$-the arm reward regression, and $\varepsilon$ is a conditionally independent subgaussian noise~\cite{li_contextual-bandit_2010,chu_contextual_2011}.

Originally~\cite{li_contextual-bandit_2010} used Upper Confidence Bound (UCB) for picking actions, which selects $
    a_t \arg\max_k x_t^\top \hat{\theta}_k
        + \mathrm{ucb}_k\bigl(x_t; \mathcal{H}_{<t}\bigr)
$, where $\hat{\theta}_k$ is the current $k$-th arm's reward model's coefficient estimate based on $\mathcal{H}_{<t}$, and the second term represents the arm's \emph{optimism in face of uncertainty}, computed from the \emph{best admissible linear model in a highly probable region} determined from the interaction history~\cite[ch.~19]{lattimore_bandit_2020}.
The intervention described in Section \ref{sec:adaptive-item-pair-recommendations-pharmacies} is delivered as proposed in \cite{agrawal_thompson_2013}. Namely, a Bayesian approximation approach with the Normal likelihood with Gaussian-Gamma conjugate prior is used, which allows for the assignment of better-defined probabilities for the selected actions using Thompson sampling: $a_t \sim \mathbb{P}_{\theta \sim \beta_t} \bigl(
    a_t \in \arg\max_k x_t^\top \theta_k
\bigr)$, where $\beta_t$ is the current posterior belief $p(\theta \mid \mathcal{H}_{\leq t})$. Thompson Sampling is implemented as a two-step sampling procedure. Still, it is worth noting that since the affine transformation $u \mapsto x_t^\top u$ of the conditional Gaussian part of the Gaussian-Gamma distribution yields a Normal-Gamma mixture, it is possible to sample in a single step from a location-scale student-$t$.

The linearity assumption equips linear bandits with regret guarantees that are sublinear in $T$, which stem from subgaussian concentration inequalities. They present an efficient learning alternative for use cases where the optimal action is to be selected based on a small collection of variables, for example, whether to send reminders to take medication and/or the importance of treatment adherence depending on self-reported adherence (or lack thereof) and interaction with received reminders in the previous few timesteps. It is also the approach used to send the item-pair recommendations introduced in Section \ref{sec:item-pair-recommendation} with the setup and results discussed in Section \ref{sec:adaptive-item-pair-recommendations-pharmacies}. They facilitate causal inference (i.e., statistical reasoning on under which circumstances the reminders work best) and are thus very useful for adaptive experimentation (see sSction \ref{sec:experimentation}).

\subsubsection{Beyond Linear Bandits}
\label{sub:beyond_linear_bandits}

Linear bandits present substantial limitations in their representational power~\cite{riquelme_deep_2018}. Deeper feature extractors can be used, i.e., replacing the reward model $x_t^\top \theta_a$ with $\phi(x_t, a)^\top \theta$ (or other nonlinear models) and learning the relevant feature representation $\phi$ along with the interaction. Increasing the complexity of the representations obscures statistical reasoning but can be the best alternative for intervention decisions for which optimality is expected to depend in complex ways on a variety of interrelated variables. This is the case, for example, of suggesting products to order with the goal of minimizing stockouts. The best recommendations will depend in complicated ways on the evolution of the demand and availability for multiple related products on different sites, the user's in-app activity, ordering behavior, and responses to previous suggestions, as well as on seasonal and environmental factors (disease outbreaks, weather, pollen levels,\ldots).

One of the most used approaches to developing deep neural bandits is stacked neural-linear bandit: a linear bandit~\cite{li_contextual-bandit_2010,agrawal_thompson_2013} is grafted atop a deep feature extractor,~\cite{zhou_neural_2020, xu_neural_2022}.
%

For example,~\cite{nabati_online_2021} adopts the stacked neural-linear bandit approach and utilizes a small-capacity experience replay queue~\cite{fedus_revisiting_2020}, and Gaussian-Gamma conjugate prior in the linear head to mitigate catastrophic forgetting. In particular, the method used updates a lower-level deep feature extractor $\phi$ on the experience buffer (most recent $\mathcal{H}_{<t}$), then in an empirical Bayes fashion, re-fits a prior for the bandit's head using reward variance matching based on the updated representations, and, finally, applies the Bayesian update to recompute the posterior.

The bandit problem is approached as a linearized Gaussian state-space model for the context-action-reward data in \cite{duran-martin_efficient_2022}. Specifically, the observation equation for the exogenous context $x$ and arm $a$ is modeled by a neural network $g(x, a; \theta)$, with its parameters $\theta$ being the unobserved state, which follows stochastic constant dynamics. The authors approximated the posterior of $\theta$ using multivariate Gaussian distribution and suggested updating it with partial feedback observations using the extended Kalman Filter~\cite{smith_application_1962, Daum2015}. To reduce the space complexity of the posterior, they re-parameterized $\theta$ as a point in a low-dimensional linear subspace, either randomly, or inferred from the parameters' trace during a pre-training step.

\subsubsection{Restless bandits}
\label{sub:restless_bandits}

In contrast to multi-armed bandits, the restless bandit setup (RMAB) is closer to the full MDP RL formulation in that the present actions (or inactions) affect the future, making planning a more pressing concern. Subjects have a simple internal state (often a binary switch, e.g., is the patient adhering to treatment or is the drug in stock for the pharmacy) with evolution following simple intervention dynamics that can be subject-specific. The number of available interventions (e.g. follow-up call to assess adherence and remind the patient of its importance or drug stocks available for distribution) is limited, and the goal is to maximize the number of subjects with the desired internal state (in the examples, treatment adherence across patients or well stocked pharmacies of an essential drug).

The RMAB setting explicitly considers an incomplete and imperfect information situation involving overseeing a finite number of concurrent Markov Decision Processes with the express goal of optimizing cumulative reward by budgeting interventions among the processes at each decision point. 
Note the contrast with the MAB setting, wherein it is tacitly assumed that the unknown state and arm-conditional reward distributions are \emph{stationary}. Additionally, the arms in RMAB refer to the overseen processes rather than action alternatives since the most commonly considered scenario is whether to intervene or skip.

In most RMAB applications, the Markovian dynamics of the processes looked after are unknown, which adds the exploration and estimation dimension to this concurrent control problem.
Processes' model identification thus deals with counterfactual estimation since each MDP may have different state transition probability $p$, yet for each action and passive inaction, we observe only the partial states and rewards obeying $p(s', r'\mid s, a=1)$ and $p(s', r'\mid s, a=0)$, respectively.

Computing the optimal policy in RMAB is a hard combinatorial optimization problem, and most applications resort to a polynomial-time heuristic policy known as the Whittle index policy~\cite{whittle_restless_1988}. The heuristic ranks the supervised processes according to the optimal reward subsidy, which tells how much to increase the \emph{passive reward} of the MDP to make it just as rewarding as intervening. The ranking is used for the greedy selection of the arms to play. The existence of such policy requires the so-called \emph{indexability} property to be satisfied for the Markov kernel and rewards of each process, which may be challenging to ensure in practice. It guarantees that at any given state, the optimal action does not switch from skipping back to intervention while the subsidy increases.

Equitable RMABs (ERMABs) are those with policies that also simultaneously maximize some equity function on the relative differences between rewards allocated across different pre-defined groups of arms. Sequential resource allocation with equitable objectives is essential when the decisions are of sensitive nature, as are many in the realm of health care, such as the distribution of limited stocks of essential drugs. Efficient algorithms for RMAB satisfying the maximin reward (maximization of the minimum prospective reward per group) and the maximum Nash welfare (maximization of the product of all group rewards) conditions are introduced in \cite{killian2023equitable}.

As a resource allocation approach to budget-aware planning in concurrent processes, restless bandits found their application in many areas, from addressing exploration and exploitation trade-off in protecting endangered wildlife from poaching~\cite{qian_restless_2016} to radio spectrum sensing~\cite{bagheri_restless_2015}.
%
In healthcare applications, the most notable examples include
allocating call-center resources for targeted maternal health information messaging~\cite{mate_field_2022, biswas_learn_2021, nishtala_selective_2021}, 
optimal carcinoma screening in resource-constrained setting~\cite{lee_optimal_2019},
healthcare worker visitation scheduling on regional and local level~\cite{bhattacharya_restless_2018},
optimization of timing of costly behavioral health interventions~\cite{baek_policy_2023},
tuberculosis treatment adherence~\cite{killian_beyond_2021},
and mobile health clinics deployment~\cite{ou_networked_2022}.

\begin{table*}[htpb]
  
  \caption{Impact metrics across experiments. T-test refer to daily accumulated expenditure for significant days only. LMM refers to the estimated parameters of a model on weekly values where all parameters are significant (with baseline expenditure based on the 3 previous months and normalized between 0 and 1). Bandit metrics indicate the average percentage of the participants in the adaptive intervention that were sent messages across the experiment and the fraction of weeks where most users (> 50\%) were nudged. Successful recommendations are those that result in purchases at a later time during the experiment of the pair's item ordered infrequently by the user. A significance level of 90\% is used, and - indicates non-significant results. For XP3, results are given separately for participants in the adaptive and non-adaptive intervention groups when appropriate, and when relevant, the distinction between the personalized (per) and random (ran) nudges is made.}
  \label{tab:results}
  \begin{tabular}{l c c c c}
    \toprule
    &XP1&XP2&XP3 adaptive & XP3 non-adaptive \\
    \midrule
    T-test: days with significant effect  & 0\% & 57\% & 0\% & 61\%\\
    T-test: largest effect & - & 0.12  & - & 0.28 \\
    T-test: largest statistical power & - & 0.65  & - & 0.81\\
    T-test: average effect & - & 0.10  & - & 0.25 \\
    T-test: average statistical power & - & 0.55  & - & 0.72\\   
    LMM: nudged that week   & 16.82 & 14.00  & \multicolumn{2}{c}{11.31 per / 11.72 ran} \\
    LMM: baseline expenditure & 905 & 1606  &\multicolumn{2}{c}{3101}\\
    Bandit: assigned to nudge & 26\% & 70\%  & 28.6\% per / 50.1\% ran & \\
    Bandit: majority assigned to nudge & 1/8 & 8/9  & 1/7 per / 5/7 ran &\\
    Successful recommendations & 18.2\% & 22.9\% & 11.2\% (11.1\% per / 11.2\% ran) & 15.5\%\\
\end{tabular}
\end{table*}

\begin{figure}[ht!]
  \centering
  \includegraphics[width=\linewidth]{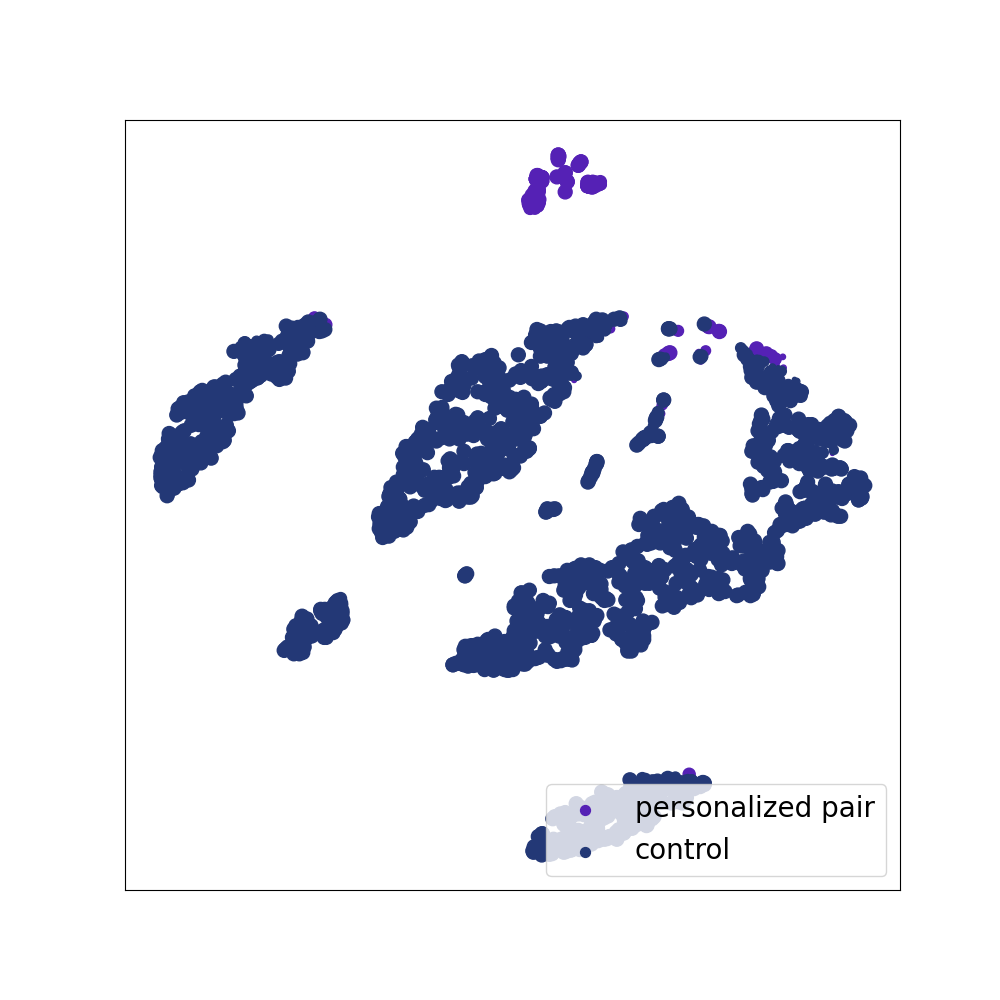}
   \caption{T-sne visualization in the contextual trait space for XP1. The figure shows the best arm for each participant at the end, with the size of the point proportional to its confidence. }
  \Description{T-sne plots for all users in XP1.}
  \label{fig:xp1-tsne-bestarm}
\end{figure}

\subsection{Experimentation}
\label{sec:experimentation}

The RL-platform allows to perform experiments in order to measure the impact of interventions. Different experimental designs are available. 
Assignment to control and the intervention strategies on trial can be fully random or adaptive, the latter using stochastic MABs (see section \ref{sec:experimentation})~\cite{Burtini2015, Yao2021, Dwivedi2022, Xiang2022}. For repeated interventions (e.g., weekly order suggestions), intra-subject assignment (i.e., assignment multiple times of the same subject throughout the experiment) is also an option to increase effective sample sizes when the impact of the intervention is expected to concentrate immediately after it is delivered, in the so-called micro-randomized trial design~\cite{klasnja_microrandomized_2015,qian_microrandomized_2022, Zhang2022} when assignment is fully randomized.

\subsection{Intervention Impact Analysis}
\label{sec:impact}

The following subsections explore methods to assess the effectiveness of interventions and measure their impact. Some methods are suitable only for randomized or adaptive designs, but all are relevant in setups using a linear bandit for adaptive interventions, which function as adaptive experiments. In these experiments a portion of the user cohort is left out as a \textit{pure control}\footnote{Note we use the terminology of \emph{control} for the no nudge arm within the adaptive intervention and of \emph{pure control} for the group of users that are not part of the adaptive intervention.} to evaluate the intervention's overall effect. This approach is used in item-pair recommendations, discussed in Section \ref{sec:adaptive-item-pair-recommendations-pharmacies}.

\subsubsection{T-tests}

Whenever the response variables considered are quantitative and normally distributed, the hypothesis test to assess whether the mean behaviour in the intervention and pure control groups is different is the Student's (when both have equal variance) or Welch's t-test (when they don't) ~\cite{fisher1935principles, box2005statistics}. For the longitudinal data collected under intervention, we apply the Welch unequal variance t-test both to the daily values of relevant metrics, as well as to the accumulated values for those metrics since the beginning of the experiment. The test statistic  $t$ for Welch's t-test is calculated as $t = (\mu_t - \mu_c)/(\sqrt{\frac{\sigma_t^2}{n_t} + \frac{\sigma_c^2}{n_c}})$ (where \( \mu_t \) and \( \mu_c \) are the sample means, \( \sigma_t^2 \) and \( \sigma_c^2 \) are the sample variances, and \( n_t \) and \( n_c \) are the sample sizes of the treatment and control groups, respectively). Under the null hypothesis, which states that the population means are equal, this test statistic follows a t-distribution with the degrees approximated by the Welch-Satterthwaite equation $\nu = (\left( \frac{\sigma_t^2}{n_t} + \frac{\sigma_c^2}{n_c} \right)^2)/(\frac{\left( \sigma_t^2/{n_t} \right)^2}{n_t - 1} + \frac{\left( \sigma_c^2/n_c \right)^2}{n_c - 1})$. By comparing the calculated $t$ value to the critical value from the t-distribution with the calculated degrees of freedom at a chosen significance level \( \alpha \) we can decide whether to reject the null hypothesis. In practice, we can compute the p-value corresponding to the test statistic. If the p-value is less than \( \alpha \), we reject the null hypothesis, indicating that there is a statistically significant difference between the two sample means. Whenever the null hypothesis is rejected, the effect size is computed as Cohen'd d ($\delta = \frac{\mu_t - mu_c}{\sigma_p}$, with \( \sigma_p \) the pooled standard deviation, calculated as $\sigma_p = \sqrt{\sigma_t^2/n_t + \sigma_c^2/n_c}$), and the associated statistical power, i.e., the likelihood that the test is correctly rejecting the null hypothesis, as $\text{Power} = P(T > t_{\alpha/2, \nu} - \delta) + P(T < -t_{\alpha/2, \nu} - \delta)$, where \( T \) follows a non-central t-distribution with non-centrality parameter \( \delta \) and degrees of freedom \( \nu \).

We also perform stratified analyses to explore heterogenous effects. That is, by dividing the study population into homogeneous subgroups, or strata, based on one or more confounding variables and performing the hypothesis testing within each stratum, we try to understand which user traits condition the size of the effect.

\begin{figure*}[ht!]
  \centering
  \includegraphics[width=0.8\linewidth]{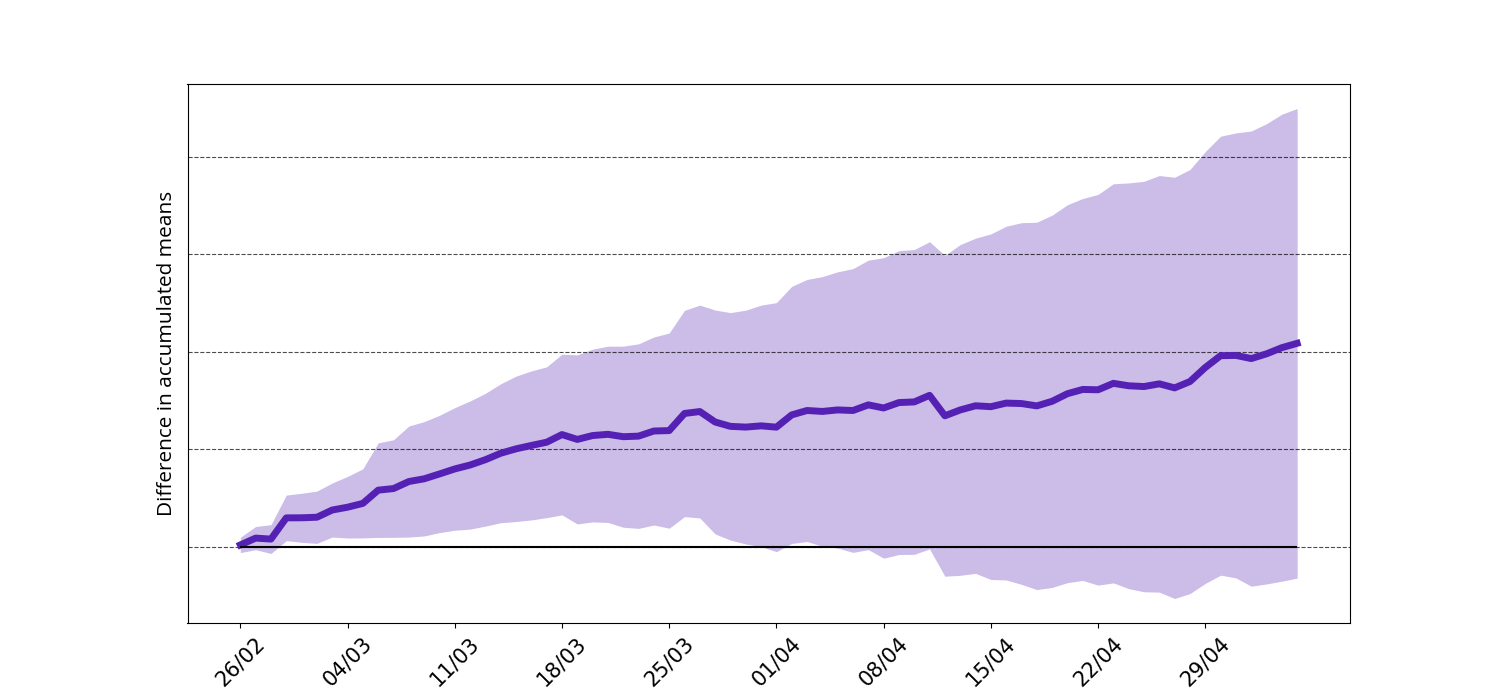}
  \caption{Difference between mean accumulated daily expenditure of users in the adaptive group vs. pure control for XP2. Confidence intervals (90 \%) associated to the t-test are shaded. The black horizontal line is drawn across 0. }
  \label{fig:xp2-rct-acc}
  \Description{T-test related plot for all users in XP2.}
\end{figure*}

\subsubsection{Linear mixed-effects models}
\label{sec:lmms}

A linear mixed effects model (LMM) extends the classical linear model by incorporating both fixed effects, which are the same across individuals, and random effects, which vary between individuals. In RCT with repeated measurements for the same subjects over time, LMMs model the within-subject correlation over time. Fixed effects are the systematic influences shared across all subjects, and random effects capture subject-specific variability~\cite{laird1982, fitzmaurice2012}. The general form of a linear mixed effects model is \( y_{ij} = \beta_0 + \beta_1 X_{ij} + b_i + \epsilon_{ij} \), where \( y_{ij} \) is the response variable for subject \( i \) at time \( j \), \( \beta_0 \) and \( \beta_1 \) are fixed effect coefficients, \( X_{ij} \) is the predictor variable for subject \( i \) at time \( j \), \( b_i \) is the random intercept for subject \( i \), assumed to be normally distributed with mean 0 and variance \( \sigma_b^2 \), and \( \epsilon_{ij} \) is the residual error term, assumed to be normally distributed with mean 0 and variance \( \sigma^2 \). 

In the context of experiments with digital interventions for an adaptive user journey, LMMs can be used for baseline value and other covariate adjustements and to model the effect with time. For example, in a setup where the interventions are messages that are sent adaptively to improve user engagement, we might specify
\(
y_{it} = \beta_0 + \beta_1 y_{i0} + \beta_2 \text{X}_{i} + \beta_3  \text{M}_{it} + \beta_4 \text{T}_{t} + \beta_5 (\text{X} \times \text{T})_{it} + u_i + \epsilon_{it}
\) 
where \( y_{it} \) is the outcome for subject \( i \) at time \( t \) and \( y_{io} \) its baseline value, \( X_i \) is 1 if the user is part of the intervention and 0 if in pure control, \( M_{it} \) indicates whether the subject was nudged at at time \( t \), \( \beta_0 \) is the intercept,  \( \beta_1 \) the baseline adjustment, \( \beta_2 \) is the fixed effect of the treatment, \( \beta_3 \) is the fixed effect o a nudge, \( \beta_4 \) is the fixed effect of time, \( \beta_5 \) is the fixed effect of the interaction between treatment and time, \( u_i \) is the random effect for subject \( i \), and \( \epsilon_{it} \) is the residual error term. The random effects \( u_i \) are assumed to be normally distributed with mean zero and variance \( \sigma_u^2 \), and the residual errors \( \epsilon_{it} \) are assumed to be normally distributed with mean zero and variance \( \sigma_\epsilon^2 \). The time varying effect at time \( t \) is  \( \beta_2 + \beta_5 \text{T}_{t} \) for users not nudged at time \( t \) and \( \beta_2 + \beta_3 + \beta_5 \text{T}_{t} \) for user who received a message at time \( t \). 
For the analysis presented in the results described in Section \ref{sec:results}, we use this model specification, and iteratively remove terms with non-significant parameters. Significance is assessed using the Wald test, which evaluates the null hypothesis of an estimated parameter or set of parameters being equal to zero. 


\subsubsection{Bandit assignment and sensitivity to context}

The analysis of data from experiments with adaptive designs needs less stress on statistical inference as the learning on whether there is evidence of impact for different values of the context is carried out by the bandit algorithm as the experiment runs. Within the adaptive intervention, we track the evolution of the average reward per arm and of the fraction assigned to each arm. 

To understand the role of the context we estimate how large the change in the probability of picking each of the arms would be, and in which direction, when the trait increases or decreases. These sensitivities are computed as the Jacobian of the arm probability based on softmax Thompson sampling approximation. For each trait, we compute the standard deviation of the derivative across the sample and use it to soft-threshold the sensitivity before averaging it, such that the sensitivity of arm \( a \) with respect to contextual trait \( b \) is given by  $\bar{s_{ab}} = 1/n \sum_j S_{\lambda \sigma_{ab}}\bigl(z_{ab}(x_j)\bigr)$ with $\lambda = 1/5$where $S_\lambda(x)$ the soft-thresholding operator $S_\lambda(v) = v \max\{1 - \lambda/\lvert v \rvert, 0\}$ with $\lambda \geq 0$. These sensitivities are then normalized and transformed into one of the 7 categories: negligible, and small, medium or large negative or positive sensitivity for ease of interpretation. To further understand whether the context as a whole has been adequately selected, we use t-Distributed Stochastic Neighbor Embedding (t-SNE) visualizations in the contextual trait space of both the best arm and its confidence (i.e., the difference between the assignment probabilities of the best and second best arms). T-SNE is a nonlinear dimensionality reduction technique used to visualize high-dimensional data by converting similarities between data points into joint probabilities and minimizing the Kullback-Leibler divergence between these probabilities in a low-dimensional space~\cite{hinton_stochastic_2002, maaten_visualizing_2008}.

\subsubsection{Recommendation success analysis}

For all message-based nudges we also track how the users interacted with them, i.e., whether the messages were opened (by tapping on them), closed (clicking on the x symbol on their top right hand corner) or ignored (neither opened nor discarded). When the messages contain specific recommendations, we also track how many of these are then realized by the users (the recommended product purchased in this particular case), and how this depends on whether the message was opened, discarded or ignored. 

\begin{figure}[ht!]
  \centering
  \includegraphics[width=0.75\linewidth]{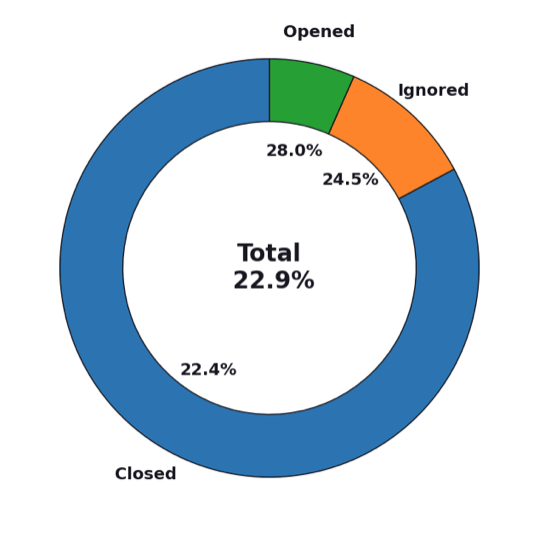}
  \caption{Breakdown by interaction (opened, closed or ignored) of messages that lead to purchase of the less frequently bought item in XP2 (22.9\% of all messages), including the percentage of each interaction these represent.}
  \Description{Recommendation success related pie chart for XP2.}
  \label{fig:xp2-success}
\end{figure}

\subsubsection{Qualitative interviews}

Finally, whenever possible, we conduct interviews with some of the users subjected to the interventions. In the case described in Section \ref{sec:adaptive-item-pair-recommendations-pharmacies} eight of the users who received the recommendations during the first experiment (see Section \ref{sec:experiments}) were interviewed, with questions concerning the user's rol, general purchasing habits, and relation to the recommended item before bringing up the messages. They were then asked explicitly for feedback on the recommendations received.

\section{Adaptive Item Pair Recommendations}
\label{sec:adaptive-item-pair-recommendations-pharmacies}

The adaptive item pair recommendations that are being used in SwipeRx and for which several experiments have already been run are used to illustrate the potential of the framework described in this paper. 

\begin{figure}
\includegraphics[width=0.9\linewidth]{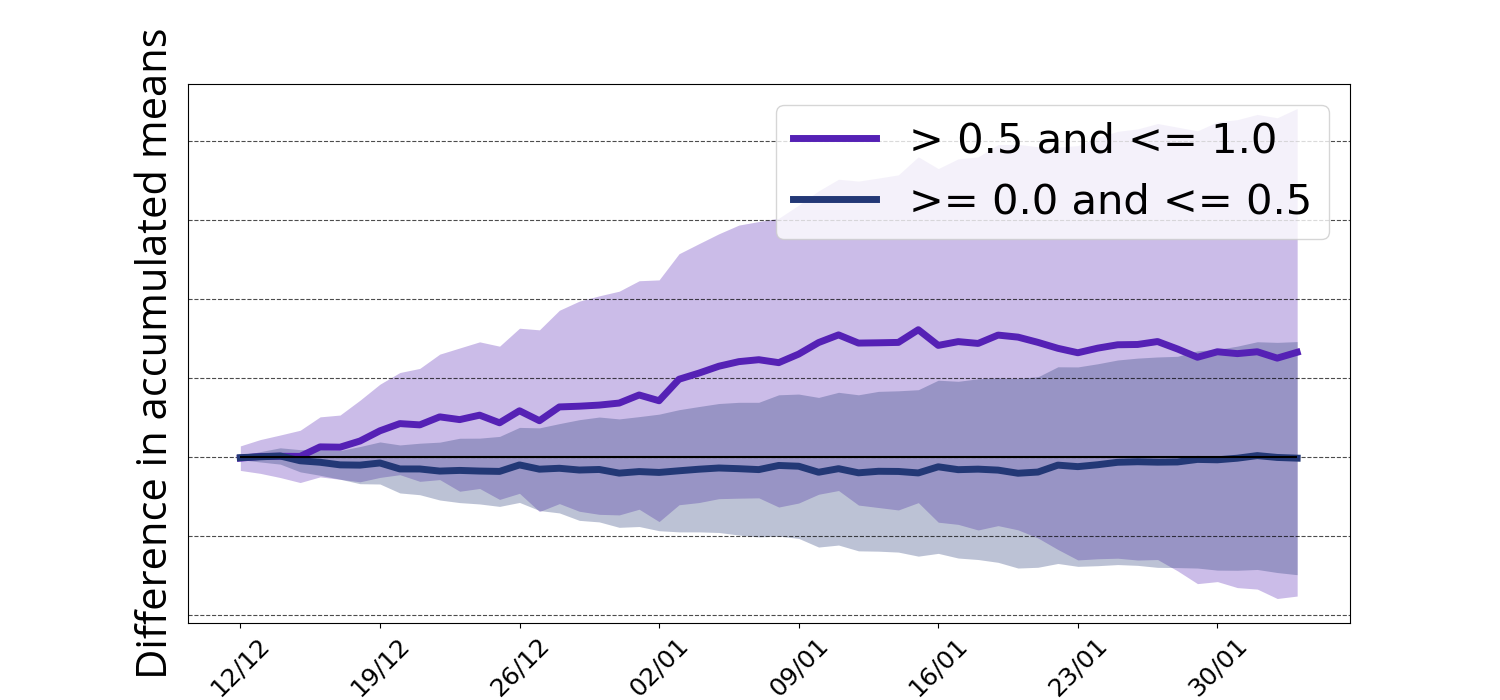}\\
  \includegraphics[width=0.9\linewidth]{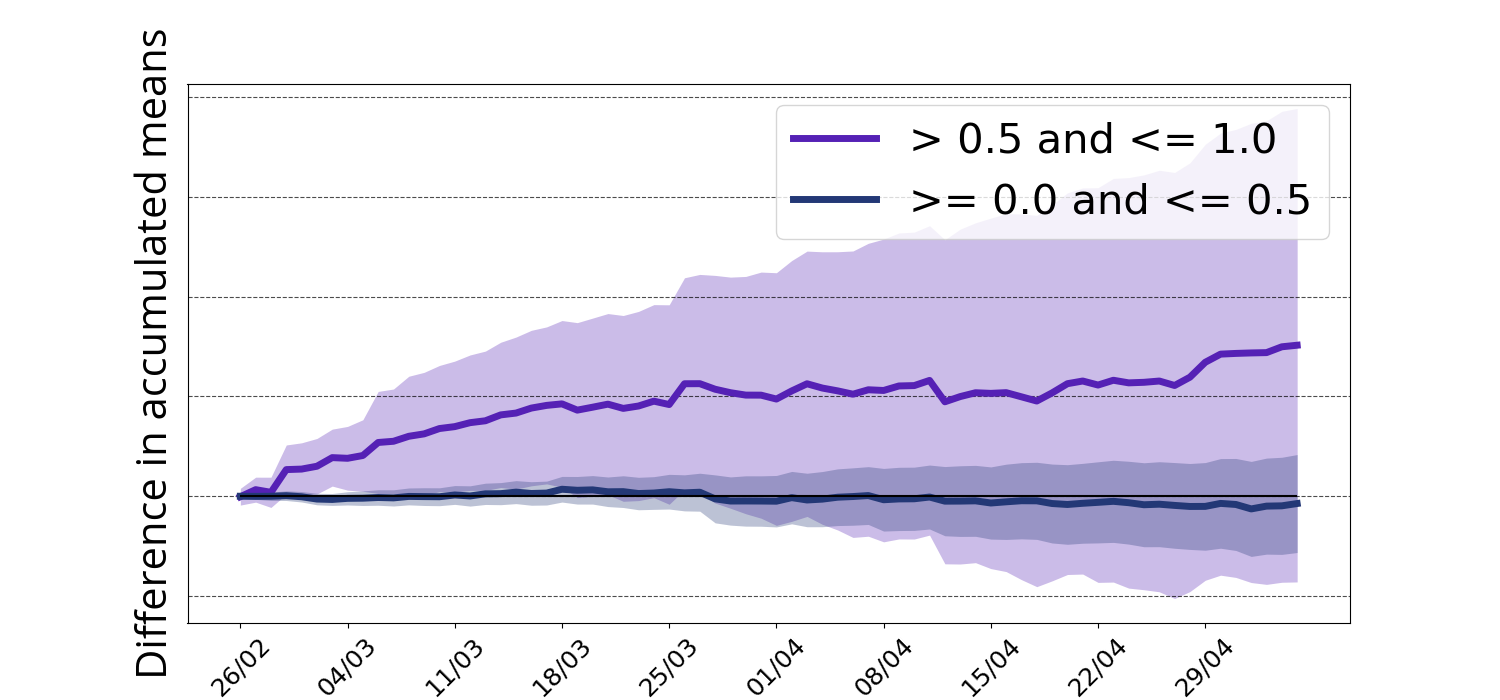}
 
  \caption{Difference between mean daily accumulated expenditure of users in the adaptive intervention vs. pure control for XP1 (top) and XP2 (bottom) stratified by the spending percentile. Confidence intervals (90 \%) are shaded. }
  \Description{Stratified in spending percentile of t-test plots for daily accumulated difference in means indicates a higher effect on high spenders.}
  \label{fig:strat-spending-rct-acc}
\end{figure}

\subsection{Intervention Setup and Experiments}

\label{sec:experiments}

The intervention involves sending weekly in-app messages with item-pair recommendations (see Section \ref{sec:item-pair-recommendation}), which appear as popups (or on the homepage when they open the app if users were inactive when they were sent). The messages, in Bahasa, state: "Pharmacies in your area typically purchase A and B. Click here to order now!" Top 20\% spenders were excluded from the initial experiments, targeting users in Indonesia with Bahasa language settings, linked to pharmacies with one or two users, and who logged in regularly (at least once in the previous 40 days and at least once a week on average on the previous two months). A Gaussian-Gamma linear bandit model was used in the adaptive arm, with the (log transformed) total expenditure over the next week as the reward. Three experiments were conducted, each lasting 8-10 weeks. The context always included the normalized days since the last nudge, some measure of purchasing frequency (over the last 90 days for the first experiment and over the last 30 days for the second and the third) and of baseline expenditure (over the last 90 days for the first and third experiment and on the ongoing month for the second), with different experiments incorporating different additional traits.

In the first experiment (XP1), messages were initially sent on Tuesdays and later switched (on week 5) to Mondays (in both cases at 6 am local time) to align the nudging with the day of the week when the largest fraction of purchases take place. The second experiment (XP2) used a smaller sample size, since all users in the intervention in XP1 were excluded. XP1 also included region as an additional context while XP2 incorporated login frequency (in the last 60 days) and in-app time (in the last 30 days). For XP1 and XP2, around 30\% and 40\%of participants assigned to pure control respectively.

The third experiment (XP3) added a non-adaptive (i.e., where subjects were nudged every Monday) treatment group with random item-pair recommendations (the same message as the personalized ones but with both items selected at random) and included an additional arm in the adaptive intervention with the same random recommendations. Participants from XP1 were included again, but not from XP2. XP3 had a participant split of 5\% non-adaptive, 60\% adaptive, and 35\% pure control, with context similar to XP2 but differing in the baseline expenditure periods and excluding 40\% (instead of 20\%) of top spenders. When discussing the results below, we will refer to the adaptive or MAB experimental group interchangeably, and similarly for non-adaptive and AB (as in AB test).

\subsection{Results}
\label{sec:results}

All impact metrics considered across the experiments are consistent with a significant uplift in expenditure due to the intervention. The impact is most prominent when the recommendations are a novelty, with the effect decaying as the experiments unfold. Users previously exposed to the same nudging in an earlier experiment are less likely to respond positively. There is, however, no indication of this fatigue turning the effect into negative over the periods considered. There is also some evidence that higher spenders react better to the recommendations. While the most significant part of the effect is concentrated in the week a nudge is received, there are indications of delayed effects. Table \ref{tab:results} compares a few impact metrics across the experiments for the different monitoring angles described in Section \ref{sec:impact}. Additional details are collected for the interested reader in the appendices \ref{app:xp1} to \ref{app:xp3}, which include, for each experiment, whenever not included in the main paper, plots of the daily and accumulated expenditures with associated t-test confidence intervals, the bandit's rewards, arm assignment and t-sne visualization of best arm and confidence, and the breakdown by interaction of successful recommendations; and tables with the bandit's sensitivity to its context and the LMM estimated parameters, both for the full expression presented in Section \ref{sec:lmms} and for the final one with all terms significant (and used in Table \ref{tab:results}).

The beginning of the first experiment (XP1) suffered from the lack of alignment between the nudging and the day of the week with higher purchasing engagement and from its start in the middle of the anomalous Christmas period. However, it shows an increase in the average basket size, even if not statistically significant according to the t-test, particularly after Christmas and concurrently with the change in nudge scheduling to Monday. The bandit here learns to progressively send fewer nudges as the experiment unfolds (contrary to what happens in the other two experiments). This could be hindering the significance of the t-test analysis (many users in treatment are not receiving recommendations most of the time), the fact that the LMM estimates a positive effect, that over 20\% of the recommended items end up being purchased during the experiment (and well over 50\% of those being products that had never been purchased before), and the sensitivity analysis, all indicate the bandit is doing a reasonable job in sending nudges to users that will benefit from them. The LMM is unable to simultaneously estimate as significant the effect of being in the intervention and of receiving a recommendation in that week, probably due to the relative lack of success of the recommendations sent on Tuesdays during the first few weeks. However, it is possible to estimate either of them significantly separately, with the effect of each nudge being estimated with much higher certainty and impact. Figure \ref{fig:xp1-tsne-bestarm} shows the t-sne visualization of the best arm, with the size of the points indicating the confidence (which is also plotted independently in Figure \ref{fig:xp1-tsne-confidence} in Appendix \ref{app:xp1}). Users receiving the nudges are mostly localized in one area, and most users have high confidence regardless of their best arm. There is a small sensitivity to some regions, a medium positive sensitivity to how long ago the user was nudged, and a negative one to the purchasing frequency.

The second experiment (XP2) is the one to more clearly indicate evidence of a significant positive impact, even if the sample sizes were significantly lower than the ones used for XP1. The bandit assigned by chance a larger part of the higher spenders to the nudge arm on the fourth week and was subsequently misled into thinking the interventions were more broadly effective than they were, which affected the sensitivities too (note that the LMMs also highlight the normalized baseline expenditure as the most important predictor of spending throughout all experiments). As a result, it was also challenging for the LMM to distinguish between intervention and nudge that week, accounting for the lack of simultaneous significance of both parameters while still clearly estimating a significant effect of the intervention. On the contrary, it made it easier to detect the impact with the t-test, as many users in the adaptive arm received messages throughout the experiment. Figure \ref{fig:xp2-rct-acc} shows the daily difference in accumulated expenditures (since the beginning of the experiment). The fact that the bandit might have assigned suboptimally more users to the recommendation probably made the effect more concentrated but less long-lasting than if it had been running over extended periods.

Figure \ref{fig:xp2-success} shows the breakdown of the 22.9\% of messages for which the recommended item was purchased according to how the user interacted with them (also for XP2). We see across the experiments that, even if opened messages have a higher likelihood of ending in purchase, the difference is not large, with most of the orders actually coming from closed messages in absolute numbers (as that is the most frequent way of interacting with the messages). The qualitative interviews indicated that, indeed, many users considered the messages informative and decided to close them to act on them later.

Both the participation of lower spenders on average (due to the exclusion of the top 40\%) and the inclusion of XP1 participants (for which the novelty effect had subsided) appear to have hindered XP3 results. The t-test still shows a significant difference for the non adaptive group, but this is also clearly the case before the intervention, indicating a very biased (small) sample. Still, random nudges also have a clear positive impact, both according to the LMM, the bandit's decisions in the adaptive arm, and the recommendation success analysis. This could be because they are cheaper on average (particularly as XP3 has the lowest spenders, many of them on tighter budgets). Their randomness could be doing an excellent job in helping pharmacists discover products they were unaware were available through SwipeRx. Interestingly, the adaptive experiment shows larger magnitude sensitivities on average for the random arm than the rest, indicating part of the effect may be due to the bandit learning effectively which users would benefit more from these random recommendations, who are users that very frequently log in and purchase with a slight tendency to small expenditures. 

Figure \ref{fig:strat-spending-rct-acc} shows the stratified analysis comparing the top 50\% to the bottom 50\% spenders for XP1 and XP2, which could be indicating the former benefits more from the recommendations.
 XP3 is not shown but indicates no clear effect on either of the strata for the adaptive group and a clear increasingly significant difference for the top 50\% for the non-adaptive that can be attributed to the biased sample selection discussed above. None of the bandits, however, estimated non-negligible sensitivity to the baseline expenditure, as they produce higher than average rewards regardless of the group in which they are placed. This fact could also explain part of the success of the random recommendations in the AB group, which is higher at 15.5\%  than the personalized and random ones in the adaptive group.

If a user has been made aware that products they typically purchase elsewhere are also available through SwipeRx, they might act on that knowledge later down the line when they need to restock it. The indication that this delayed effects exist comes mainly from the recommendation success analysis, which shows that a small fraction of these purchases occur weeks after the in-app message. Additionally, while not significant (and not shown here), the trends in differences in the accumulated expenditure are at least compatible with this observation. 

\section{Summary and Conclusions}

We have introduced a framework enabling message- and content-based interventions in healthcare-related digital tools through integration with an RL platform. We have also described how it could enrich the user experience and provide an adaptive user journey. To illustrate this approach, we have discussed the results of a series of initial experiments with personalized and adaptive item-pair recommendations delivered through in-app messages to customers of SwipeRx, the largest all-in-one app for pharmacies in Southeast Asia. The significant increase in basket size measured highlights the potential of this framework, which is flexible by design to enable integration with already existing tools to scale its impact.

\begin{acks}
The authors want to thank Susan Murphy for insightful discussions. This work was supported, in whole or in part, by the Bill \& Melinda Gates Foundation INV-060956. Under the grant conditions of the Foundation, a Creative Commons Attribution 4.0 Generic License has been assigned to the Author Accepted Manuscript version that might arise from this submission.
\end{acks}

\bibliographystyle{ACM-Reference-Format}
\bibliography{main}

\appendix

\section{XP1 Analysis}
\label{app:xp1}

\subsection{T-tests}

\begin{figure*}[ht!]
  \centering
  \includegraphics[width=\linewidth]{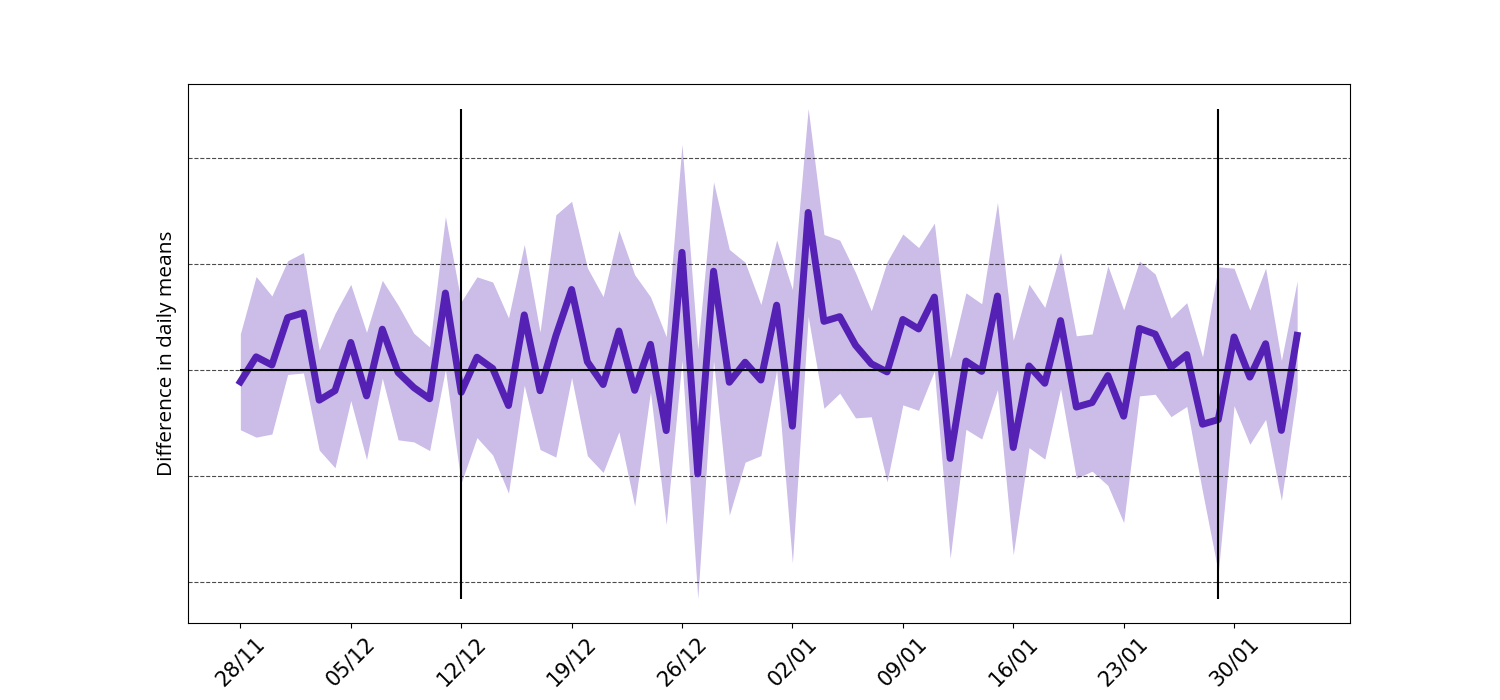}\\
  \includegraphics[width=\linewidth]{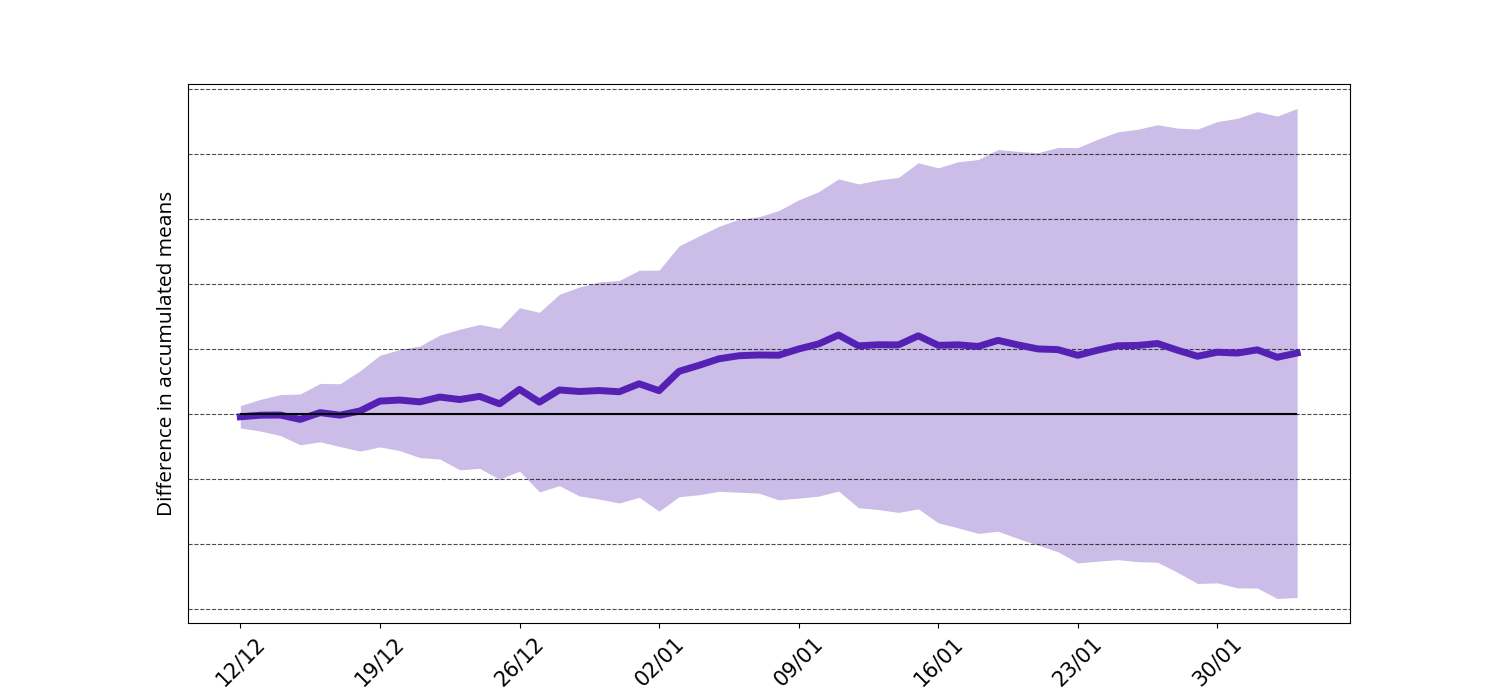}
  \caption{Difference between mean daily (top) and accumulated daily (bottom) expenditure between all users in the adaptive arm vs. control in the randomized control trial XP1. Confidence intervals (90 \%) are shaded. The black horizontal line is drawn across 0, and vertical lines represent the intervention period's beginning and end. }
  \Description{T-test related plot for all users in XP1.}
  \label{fig:xp1-rct}
\end{figure*}

Figure \ref{fig:xp1-rct} shows the mean daily expenditure difference and the mean daily accumulated difference during experiment XP1 for all users. 

\subsection{Bandit Assignment and Sensitivity}

\begin{figure*}[htpb]
  \centering
  \includegraphics[width=\linewidth]{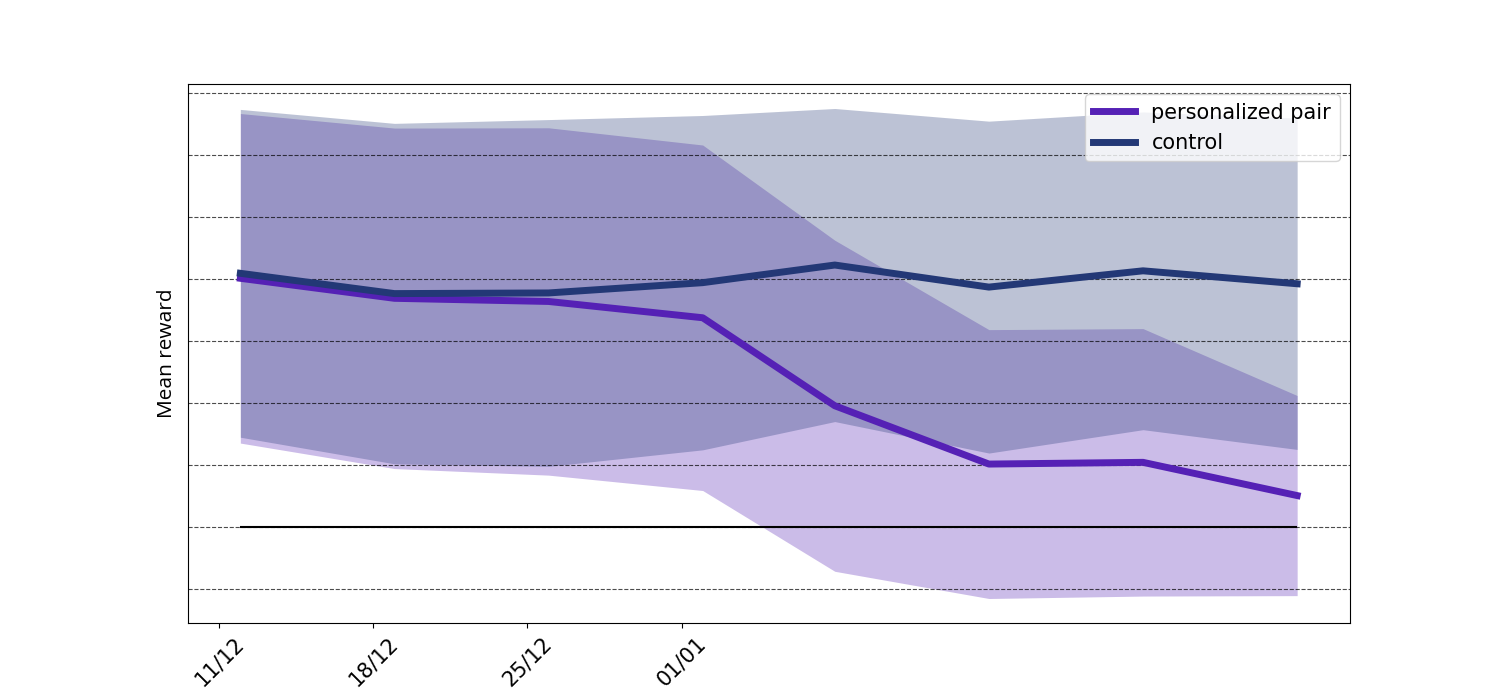}
  \caption{Mean reward (log transformed expenditure in the 6 days following the in-app message) for XP1's adaptive intervention arms. Shaded area indicates one standard deviation and the black horizontal line is drawn across 0.}
  \label{fig:xp1-rewards}
  \Description{Reward plot for XP1.}
\end{figure*}

\begin{figure*}[htpb]
  \centering
  \includegraphics[width=\linewidth]{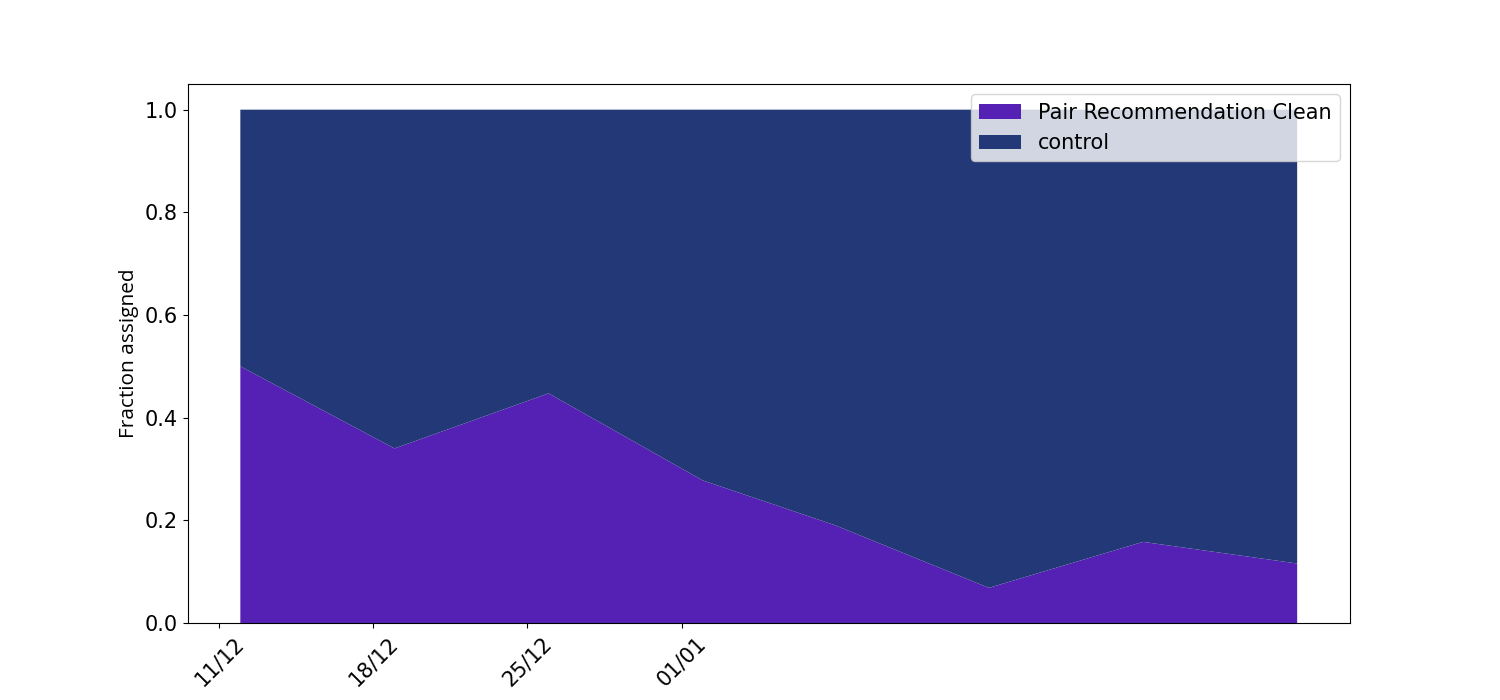}
  \caption{Fraction of participants assigned to each of XP1's adaptive intervention arms at each decision point.}
  \label{fig:xp1-split}
  \Description{Arm proportions plot for XP1.}
\end{figure*}

Figure \ref{fig:xp1-rewards} shows the evolution of the average reward per arm and decision point for XP1's adaptive intervention's arms, while Figure \ref{fig:xp1-split} the resulting proportions of assigned participants to each of the arms.

Table \ref{tab:xp1-sensitivity} contains the sensitivity classification of the item pair recommendation arm to each of the contextual traits for XP1. As there are only two arms, sensitivity of the control arm is the same in magnitude and opposite sign.

\begin{table}[htpb]
  \caption{Sensitivity to context XP1. Sensitivity categories are shown as follows: {\LARGE -} large negative, - medium negative, {\tiny -} small negative, empty negligible, {\tiny +} small positive, + medium positive, {\LARGE +} large positive. }
  \label{tab:xp1-sensitivity}
  \begin{tabular}{p{5cm} p{2cm}}
    \toprule
    &Item-pair\\
      \toprule
    Expenditure previous 3 months &  \\
    Days since last nudge & + \\
    Days with order previous 3 months & \LARGE{-} \\
    Bali region &  {\tiny +}\\
    Banten region & \\
    DI Yogyakarta region & \\
    DKI Jakarta region & {\tiny +}\\
    Jawa Barat region & \\
    Jawa Tengah region & \\
    Jawa Timur region & {\tiny -} \\
    NAD region & \\
    Surabaya region &  \\
    \midrule

\end{tabular}
\end{table}

\begin{figure}[htpb]
  \centering
  \includegraphics[width=\linewidth]{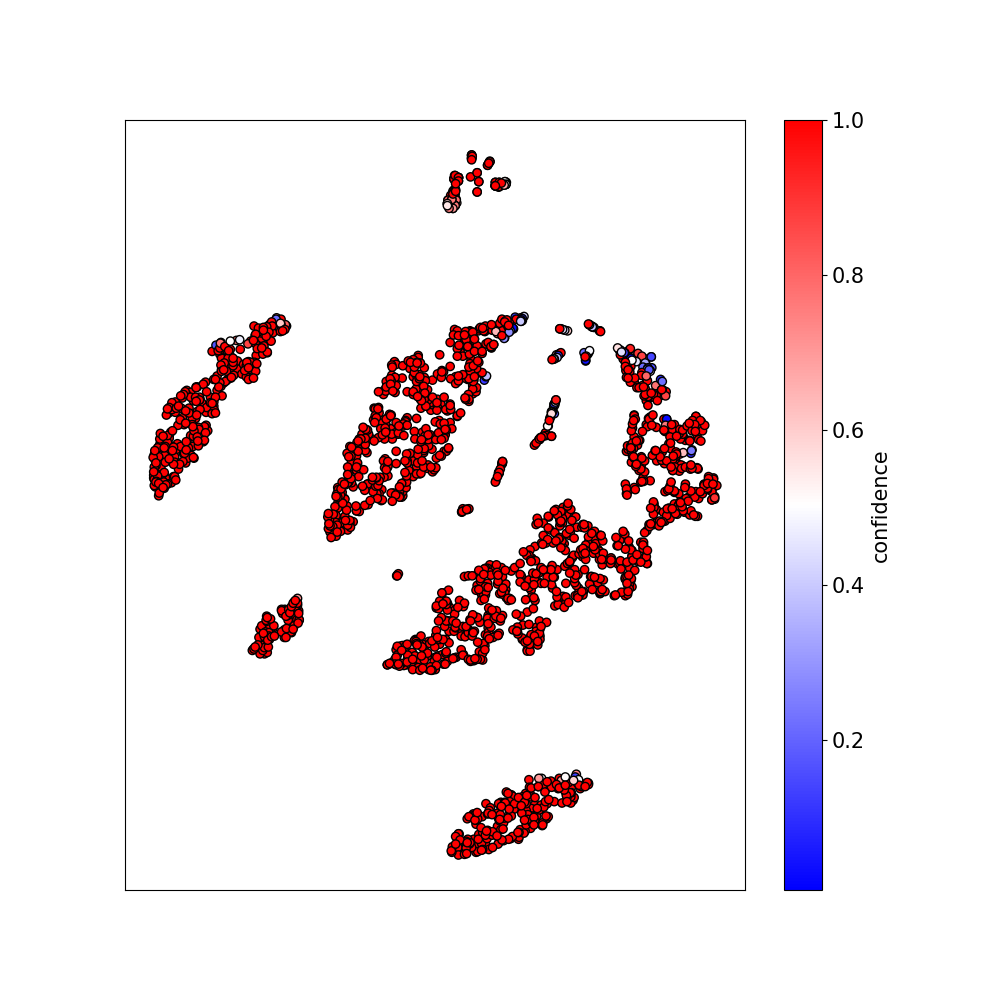}
  \caption{T-sne visualization in the contextual trait space for XP1. The figure depicts its confidence (difference between the probabilities of the best arm and the second best arm) on the last decision point. The equivalent best arm plot can be found in \ref{fig:xp1-tsne-bestarm}. }
  \Description{T-sne plots for all users in XP1.}
  \label{fig:xp1-tsne-confidence}
\end{figure}

Figure \ref{fig:xp1-tsne-bestarm} shows the t-sne visualization in the contextual trait space of the best arm and Figure \ref{fig:xp1-tsne-confidence} and of its confidence computed as the difference in the probabilities of picking the best arm and second best arm (right) for XP1.

\subsection{LMM estimation}

Table \ref{tab:xp1-lmm-full} contains the estimated parameters, standard errors and p-value of the term's Wald test for the full LMM described in Section \ref{sec:lmms} for XP1, while table \ref{tab:xp1-lmm} is the equivalent table for the fitted LMM where all coefficients are significant.

\begin{table}[htpb]
  \caption{Estimated parameters, standard error, and Wald's test p-value for the LMM with all terms for XP1 fitted to the weekly values.}
  \label{tab:xp1-lmm-full}
  \begin{tabular}{r c c c }
    \toprule
    &Coef. & Std.Err. & p-value\\
      \toprule
    Intercept & 6.585 & 3.073 & 0.032 \\
    Adaptive intervention & -1.965 & 4.047 & 0.627\\
    Nudged that week & 19.744 & 6.555 & 0.003 \\
    Baseline expenditure & 3099.689 & 29.745 & 0.000 \\
    Week number &  0.412 & 0.054 & 0.000 \\
    Week number in intervention & 0.254 & 0.075 & 0.001 \\
    \midrule

\end{tabular}
\end{table}

\begin{table}[htpb]
  \caption{Estimated parameters, standard error,and Wald's test p-value for the LMM with only significant terms for XP1 fitted to the weekly values.}
  \label{tab:xp1-lmm}
  \begin{tabular}{r c c c }
    \toprule
    &Coef. & Std.Err. & p-value\\
      \toprule
    Nudged that week & 16.817 & 6.443 &  0.013 \\
    Baseline expenditure & 3101.071 & 29.745 & 0.000 \\
    Week number & 0.545 & 0.038 & 0.000 \\
    \midrule

\end{tabular}
\end{table}

\subsection{Recommendation Success Analysis}

\begin{figure}[htpb]
  \centering
  \includegraphics[width=0.75\linewidth]{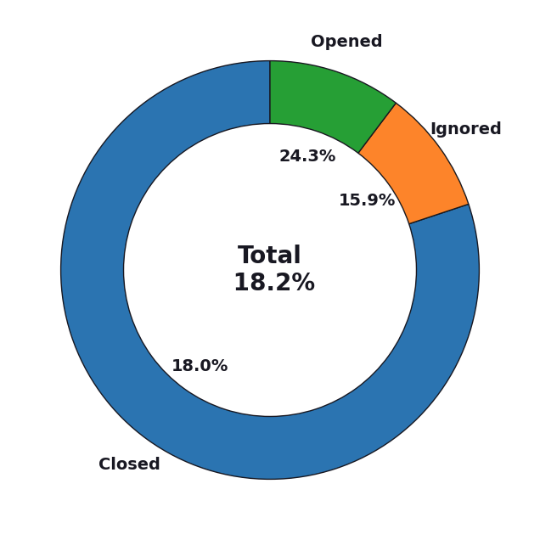}
  \caption{Breakdown by interaction (opened, closed or ignored) of messages that lead to in XP1 (18.2\% of all messages), including the percentage of each interaction these represent.}
  \Description{Recommendation success related pie chart for XP1.}
  \label{fig:xp1-success}
\end{figure}

For XP2, for 18.2\% of the messages, the item that had previously been infrequently purchased was ordered at a later time during the experiment. The breakdown by how the user had interacted with the message is reflected in Figure \ref{fig:xp1-success}. 

\subsection{Qualitative Interviews}

All pharmacists interviewed had purchased at least one recommendation, three after closing the message, two after ignoring it and another two after opening it. Several respondents recall closing the message and then searching for the product in the app and others appeared to view it as merely informative, so it seems it was unclear to many that the messages could be interacted with. A pharmacists referred their desire to be able to review the information later on as they could be busy when they received the message. Half of them recounted they found the recommendations useful and several showed interest in receiving tips when discounts are available and when products they have looked for are back again in stock.

\section{XP2 Analysis}
\label{app:xp2}

\subsection{T-tests}

\begin{figure*}[htpb]
  \centering
  \includegraphics[width=\linewidth]{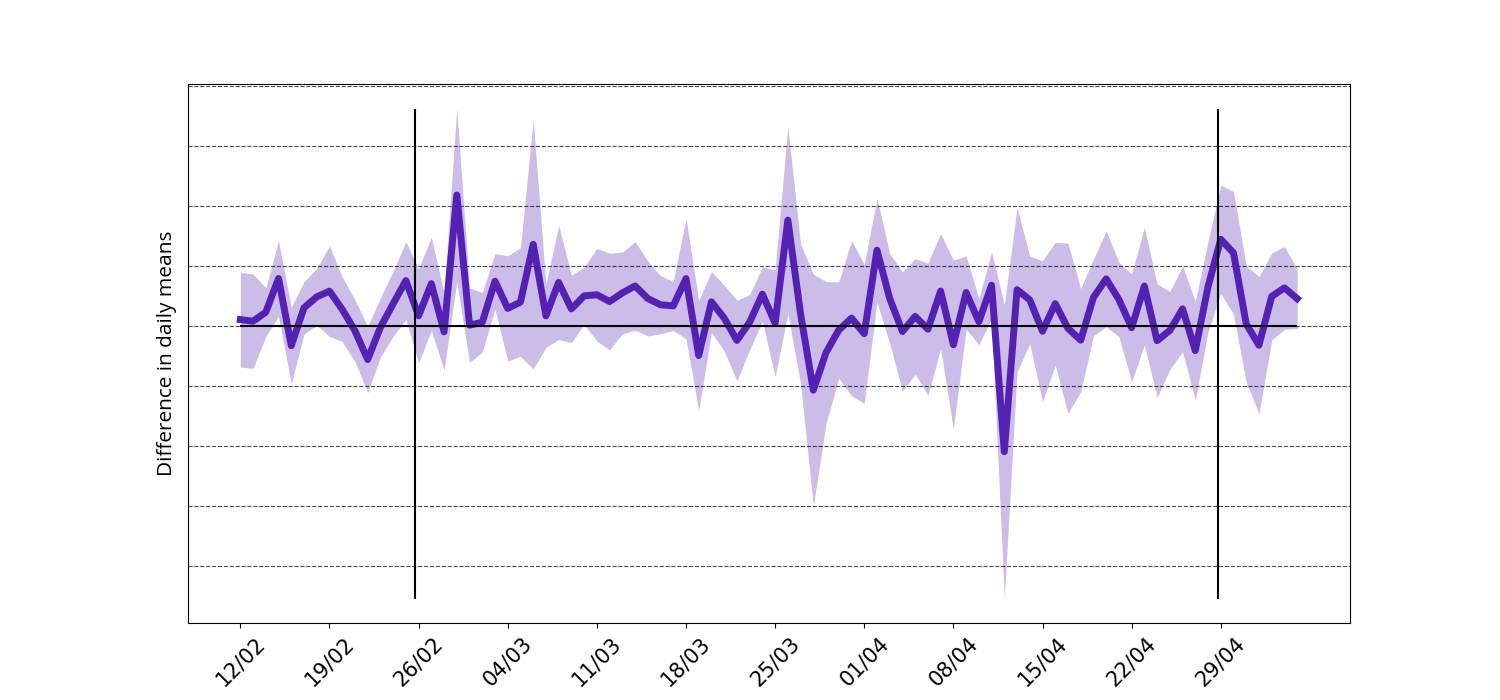}
  \caption{Difference between mean daily expenditure between all users in the adaptive arm vs. control in the randomized control trial XP2. Confidence intervals (90 \%) are shaded. The black horizontal line is drawn across 0, and vertical lines represent the intervention period's beginning and end. }
  \label{fig:xp2-rct-day}
  \Description{T-test related plots for all users in XP2.}

\end{figure*}

Figure \ref{fig:xp2-rct-day} shows the mean daily expenditure difference and the mean daily accumulated difference during experiment XP2 for all users and Figure \ref{fig:xp2-rct-acc}.

\subsection{Bandit Assignment and Sensitivity}

\begin{figure*}[htpb]
  \centering
  \includegraphics[width=\linewidth]{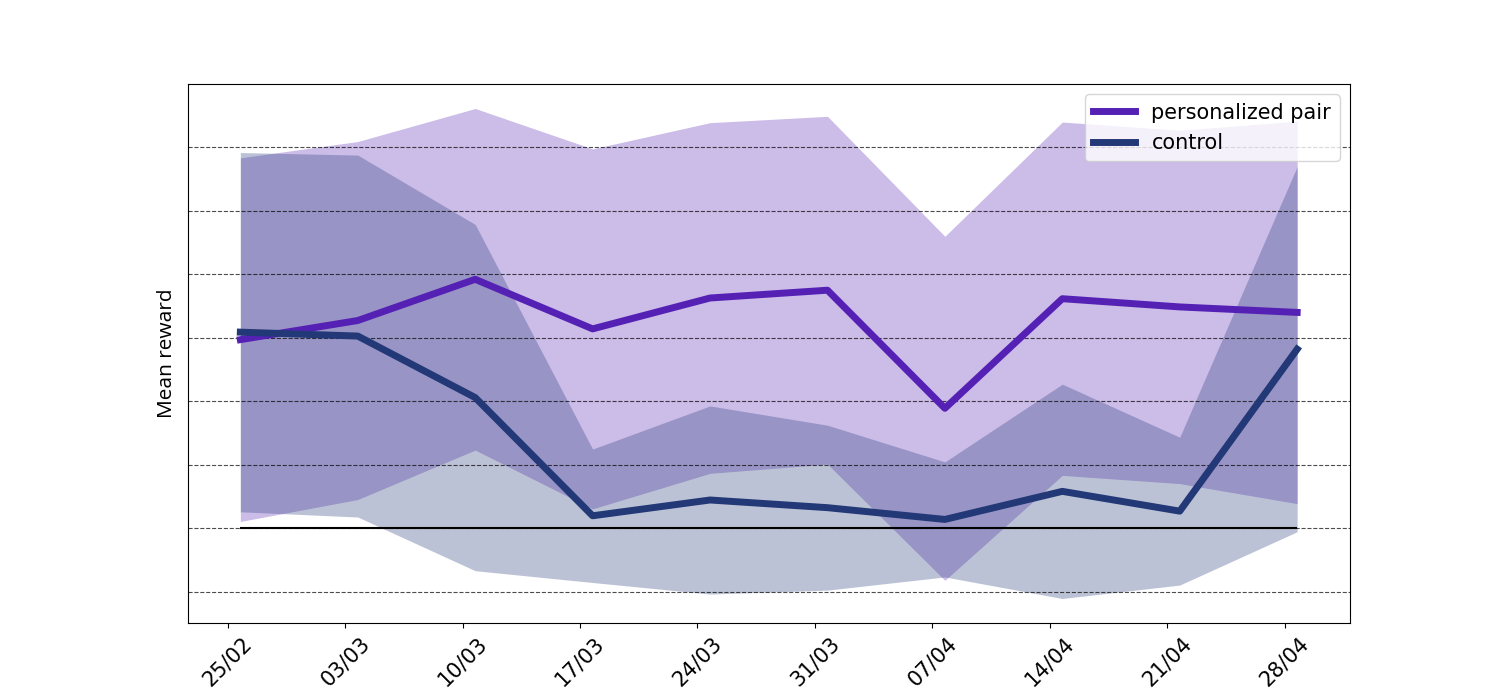}
  \caption{Mean reward (log transformed expenditure in the week following the in-app message) for XP2's adaptive intervention arms. Shaded area indicates one standard deviation and the black horizontal line is drawn across 0.}
  \label{fig:xp2-rewards}
  \Description{Reward plot for XP2.}
\end{figure*}

\begin{figure*}[htpb]
  \centering
  \includegraphics[width=\linewidth]{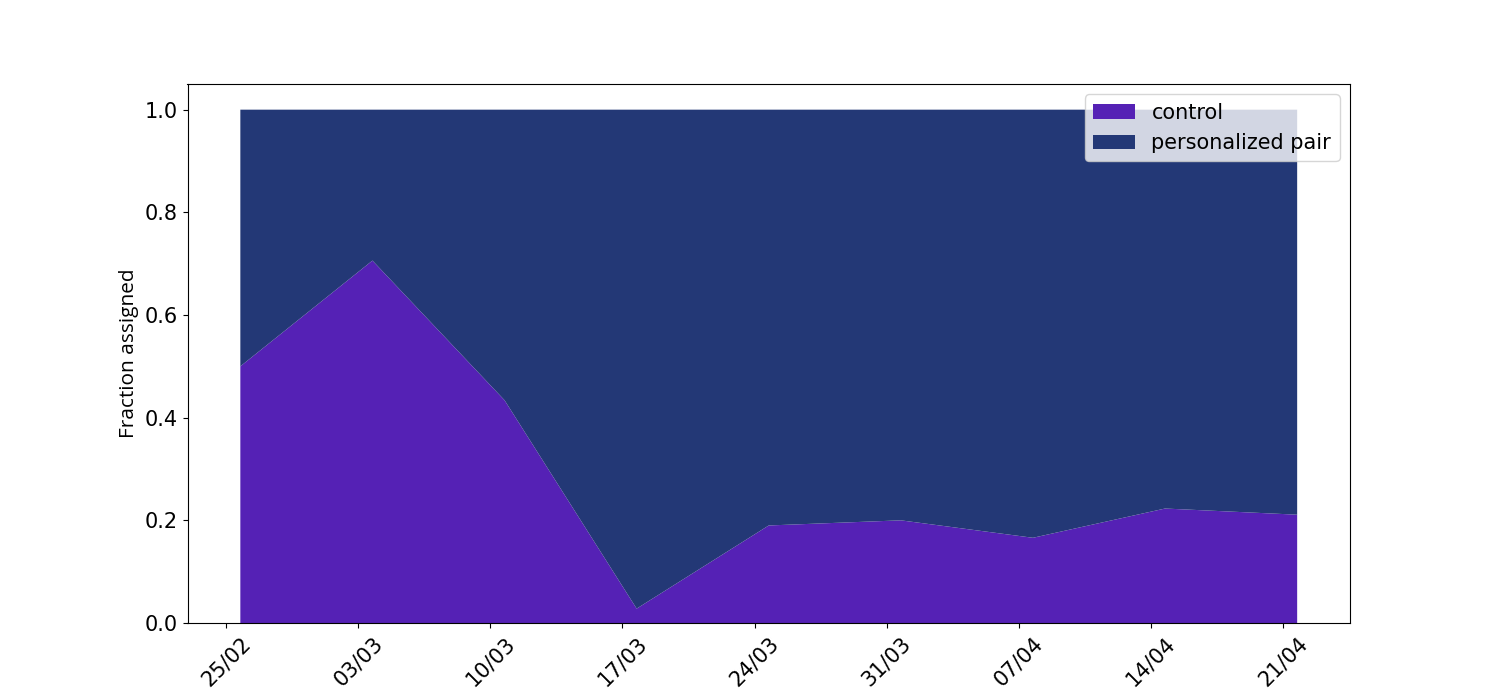}
  \caption{Fraction of participants assigned to each of XP2's adaptive intervention arms at each decision point.}
  \label{fig:xp2-split}
  \Description{Arm proportions plot for XP2.}
\end{figure*}

Figure \ref{fig:xp2-rewards} shows the evolution of the average reward per arm and decision point for XP2's adaptive intervention's arms, while Figure \ref{fig:xp2-split} the resulting proportions of assigned participants to each of the arms.

Table \ref{tab:xp2-sensitivity} contains the sensitivity classification of the item pair recommendation arm to each of the contextual traits for XP2. As there are only two arms, sensitivity of the control arm is the same in magnitude and opposite sign.

\begin{table}[htpb]
  \caption{Sensitivity to context XP2. Sensitivity categories are shown as follows: {\LARGE -} large negative, - medium negative, {\tiny -} small negative, empty negligible, {\tiny +} small positive, + medium positive, {\LARGE +} large positive. }
  \label{tab:xp2-sensitivity}
  \begin{tabular}{p{5cm} p{2cm}}
    \toprule
    &Item-pair\\
      \toprule
    Expenditure previous 30 days &  \\
    Days between logins previous 60 days & {\tiny -} \\
    Days since last nudge & - \\
    Days since first login &  \\
    Days with order previous 30 days & \LARGE{+} \\
    Opened nudges last 14 days &  \\
    In-app time last 30 days & \\
    \midrule

\end{tabular}
\end{table}

\begin{figure*}[htpb]
  \centering
  \includegraphics[width=0.45\linewidth]{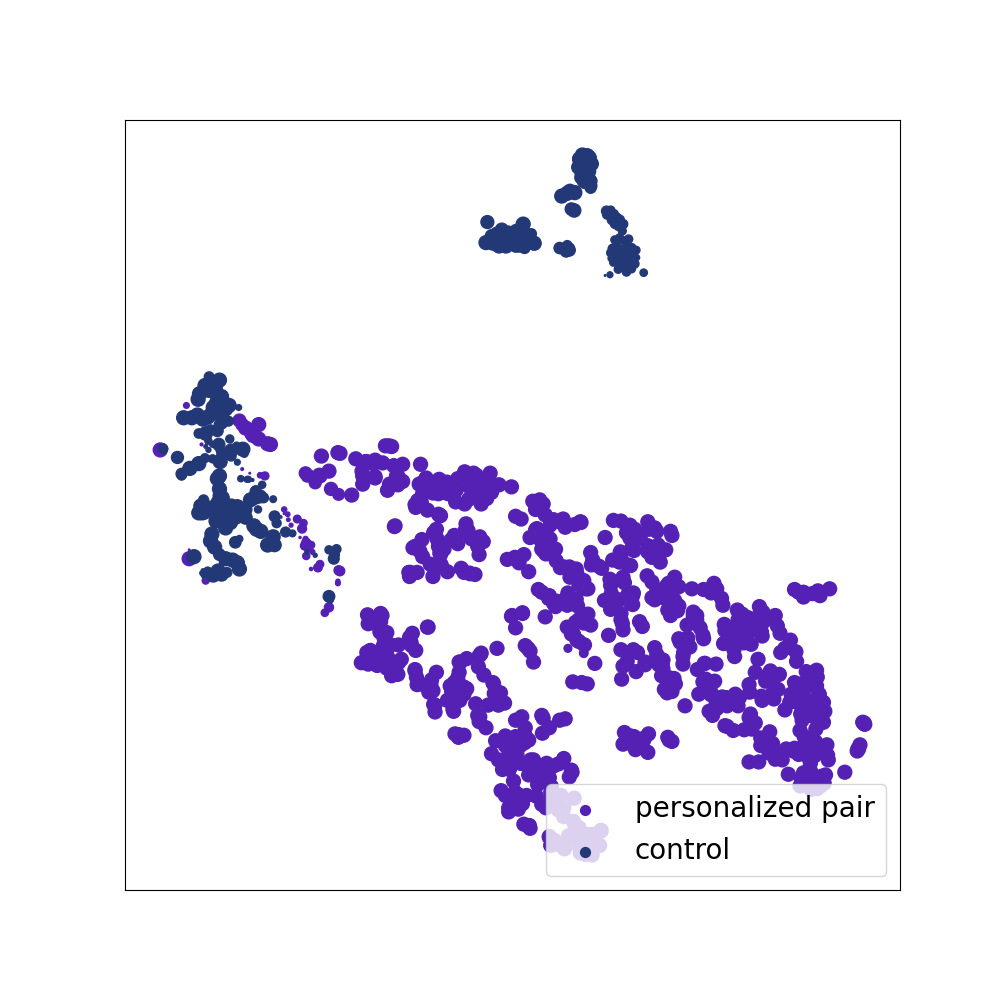}
  \includegraphics[width=0.45\linewidth]{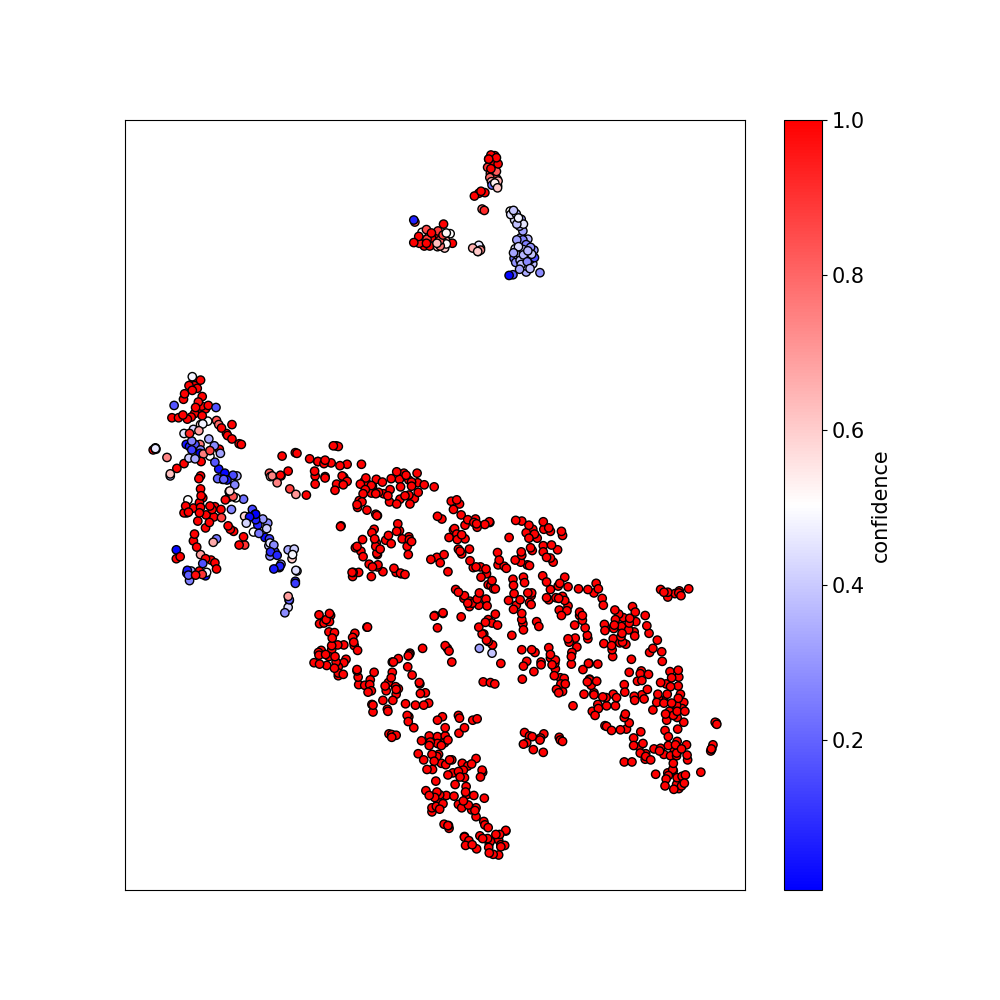}
  \caption{T-sne visualization in the contextual trait space for XP2. The figure to the left represents the best arm for each participant at the last decision point, with the size of the point proportional to its confidence (difference between the probabilities of the best arm and the second best arm), which is also plotted in the figure to the right. }
  \Description{T-sne plots for all users in XP2.}
  \label{fig:xp2-tsne}
\end{figure*}

Figure \ref{fig:xp2-tsne} shows the t-sne visualization in the contextual trait space of the best arm (left) and of its confidence computed as the difference in the probabilities of picking the best arm and second best arm (right) for XP2.

\subsection{LMM estimation}

Table \ref{tab:xp2-lmm-full} contains the estimated parameters, standard errors, and p-value of the term's Wald test for the full LMM described in Section \ref{sec:lmms} for XP2, while table \ref{tab:xp2-lmm} is the equivalent table for the fitted LMM where all coefficients are significant.

\begin{table}[htpb]
  \caption{Estimated parameters, standard error, and Wald's test p-value for the LMM with all terms for XP2 fitted to the weekly values.}
  \label{tab:xp2-lmm-full}
  \begin{tabular}{r c c c }
    \toprule
    &Coef. & Std.Err. & p-value\\
      \toprule
    Intercept & 9.331 & 5.699 & 0.102 \\
    Adaptive intervention & -2.540  & 7.295 & 0.728 \\
    Nudged that week & 15.950 & 3.891 & 0.000\\
    Baseline expenditure & 1606.137 & 32.335 & 0.000 \\
    Week number & 1.079 & 0.778 & 0.166  \\
    Week number in intervention &  -1.088 & 0.985 & 0.270 \\
    \midrule

\end{tabular}
\end{table}

\begin{table}[htpb]
  \caption{Estimated parameters, standard error and Wald's test p-value for the LMM with only significant terms for XP2 fitted to the weekly values.}
  \label{tab:xp2-lmm}
  \begin{tabular}{r c c c }
    \toprule
    &Coef. & Std.Err. & p-value\\
      \toprule
    Intercept & 10.268 & 3.308 & 0.002 \\
    Nudged that week & 13.999 & 3.580 & 0.000 \\
    Baseline expenditure & 1605.977 & 32.359 & 0.000  \\
    \midrule

\end{tabular}
\end{table}

\subsection{Recommendation Success Analysis}

For XP2, for 22.9\% of the messages, the item that had previously been infrequently purchased was ordered at a later time during the experiment. The breakdown by how the user had interacted with the message is reflected in Figure \ref{fig:xp2-success}.

\section{XP3 Analysis}
\label{app:xp3}

\subsection{T-tests}

\begin{figure*}[htpb]
  \centering
  \includegraphics[width=\linewidth]{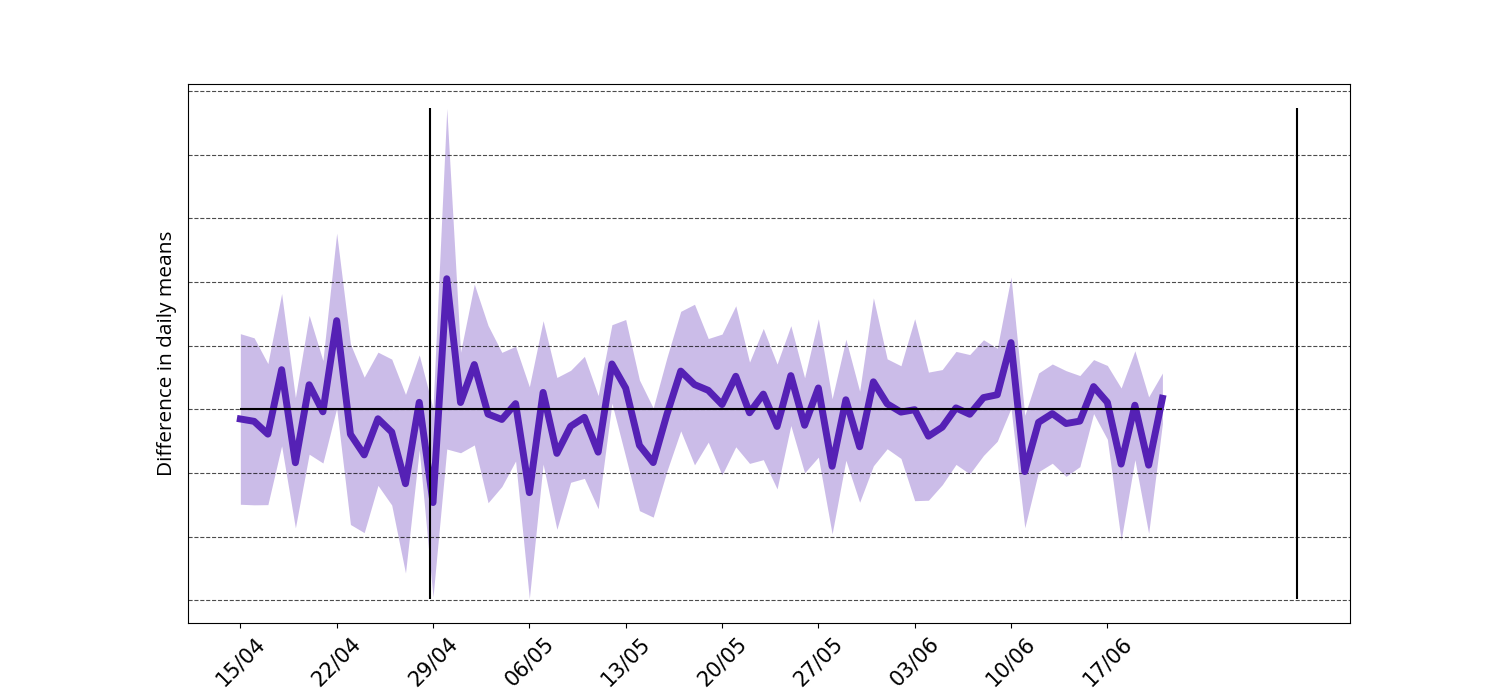}\\
  \includegraphics[width=\linewidth]{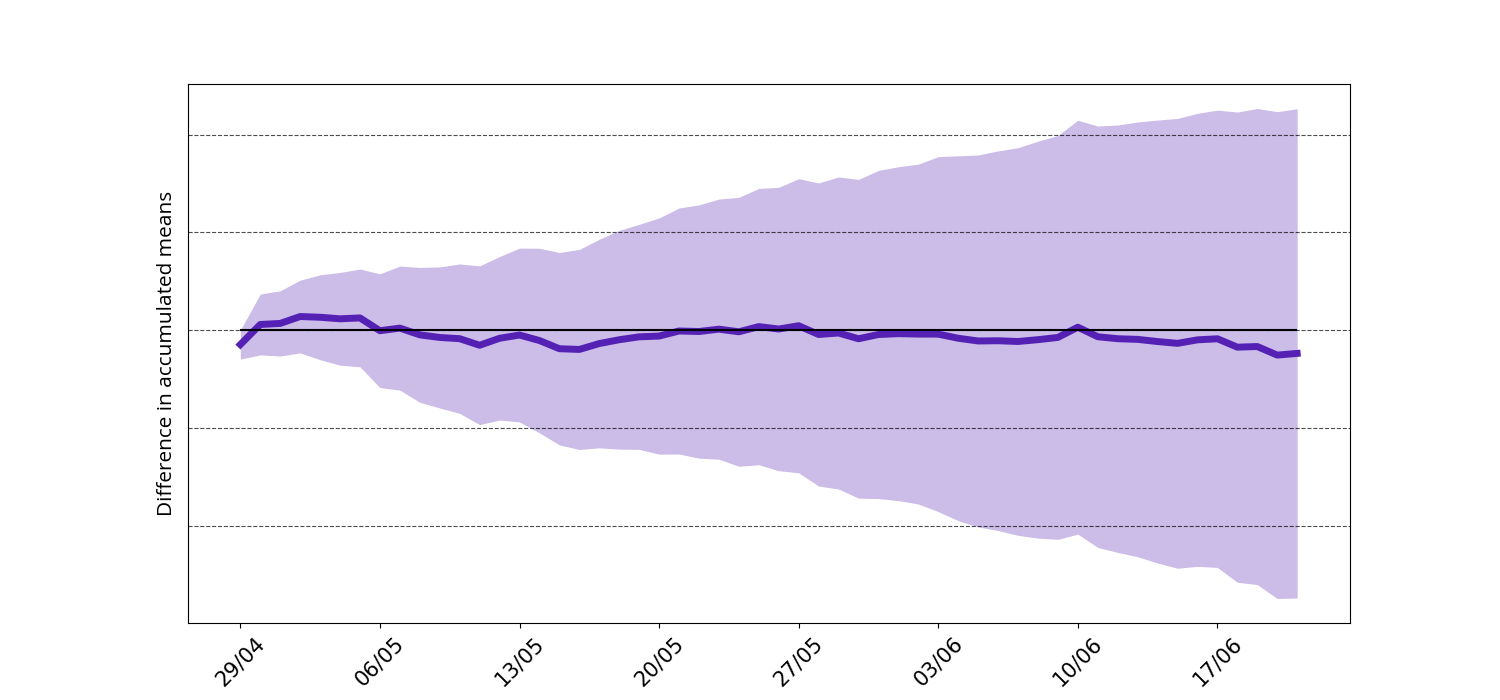}
  \caption{Difference between mean daily (top) and accumulated daily (bottom) expenditure between all users in the adaptive arm vs. control in the randomized control trial XP3. Confidence intervals (90 \%) are shaded. The black horizontal line is drawn across 0, and vertical lines represent the intervention period's beginning and end. }
    \Description{T-test related plots for all users in XP3's adaptive intervention.}

  \label{fig:xp3-rct}
\end{figure*}

\begin{figure*}[htpb]
  \centering
  \includegraphics[width=\linewidth]{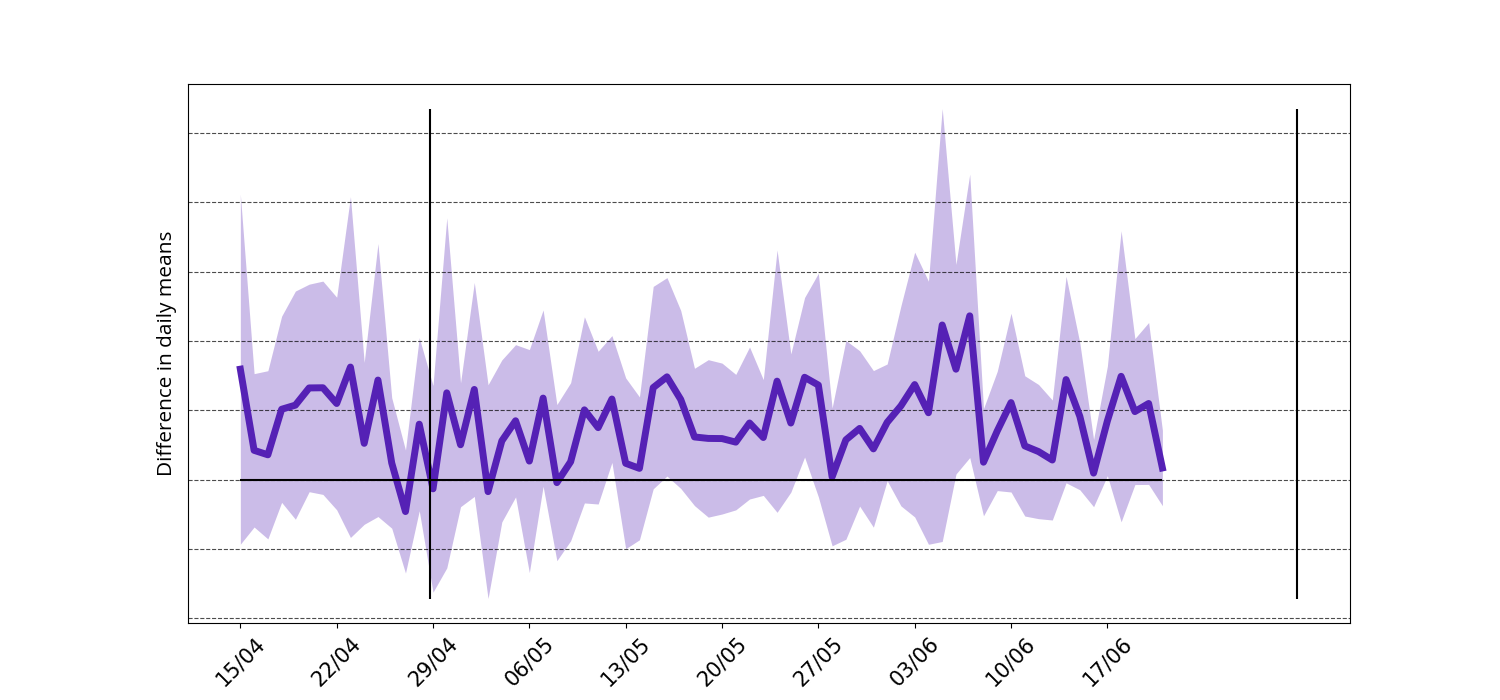}\\
  \includegraphics[width=\linewidth]{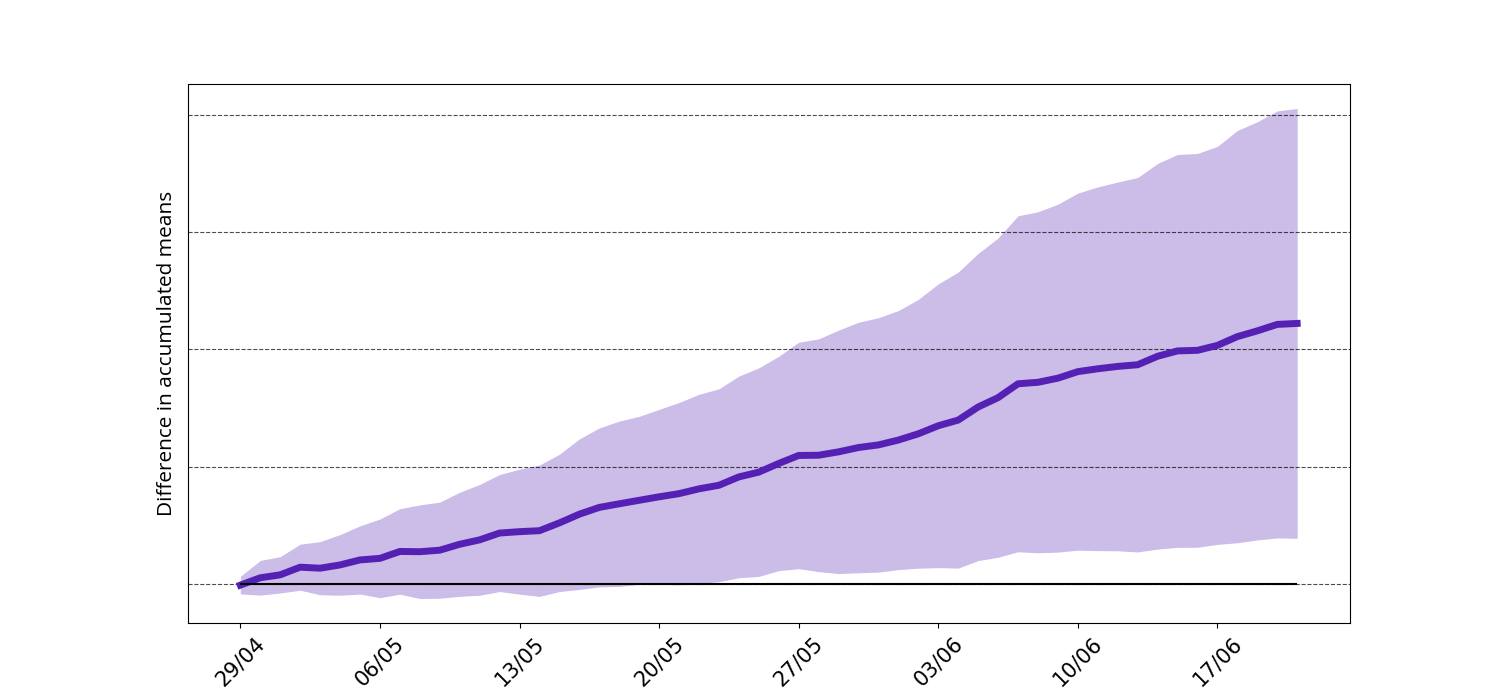}
  \caption{Difference between mean daily (top) and accumulated daily (bottom) expenditure between all users in the non-adaptive random pair recommendation arm vs. control in the randomized control trial XP3. Confidence intervals (90 \%) are shaded. The black horizontal line is drawn across 0, and vertical lines represent the intervention period's beginning and end. }
  \label{fig:xp3-rct-ab}
    \Description{T-test related plots for all users in XP3's non-adaptive random recommendation intervention.}

\end{figure*}

Figure \ref{fig:xp3-rct} shows the mean daily expenditure difference and the mean daily accumulated difference during experiment XP3 for all users for the adaptive arm as compared with pure control, while \ref{fig:xp3-rct-ab} shows the same for the AB test with non-adaptive random item-pair recommendations. 

\subsection{Bandit Assignment and Sensitivity}

\begin{figure*}[htpb]
  \centering
  \includegraphics[width=\linewidth]{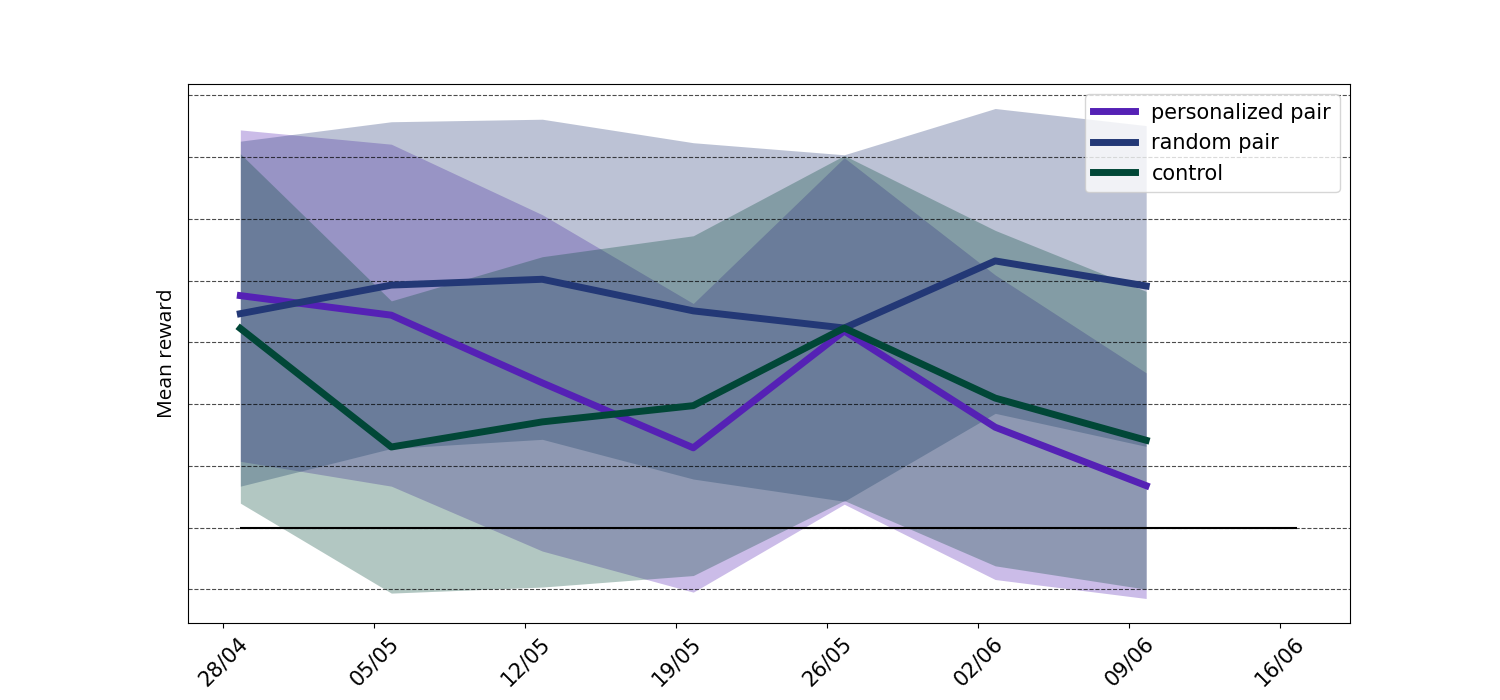}
  \caption{Mean reward (log transformed expenditure in the week following the in-app message) for XP3's adaptive intervention arms. Shaded area indicates one standard deviation and the black horizontal line is drawn across 0.}
  \label{fig:xp3-rewards}
  \Description{Reward plot for XP3.}
\end{figure*}

\begin{figure*}[htpb]
  \centering
  \includegraphics[width=\linewidth]{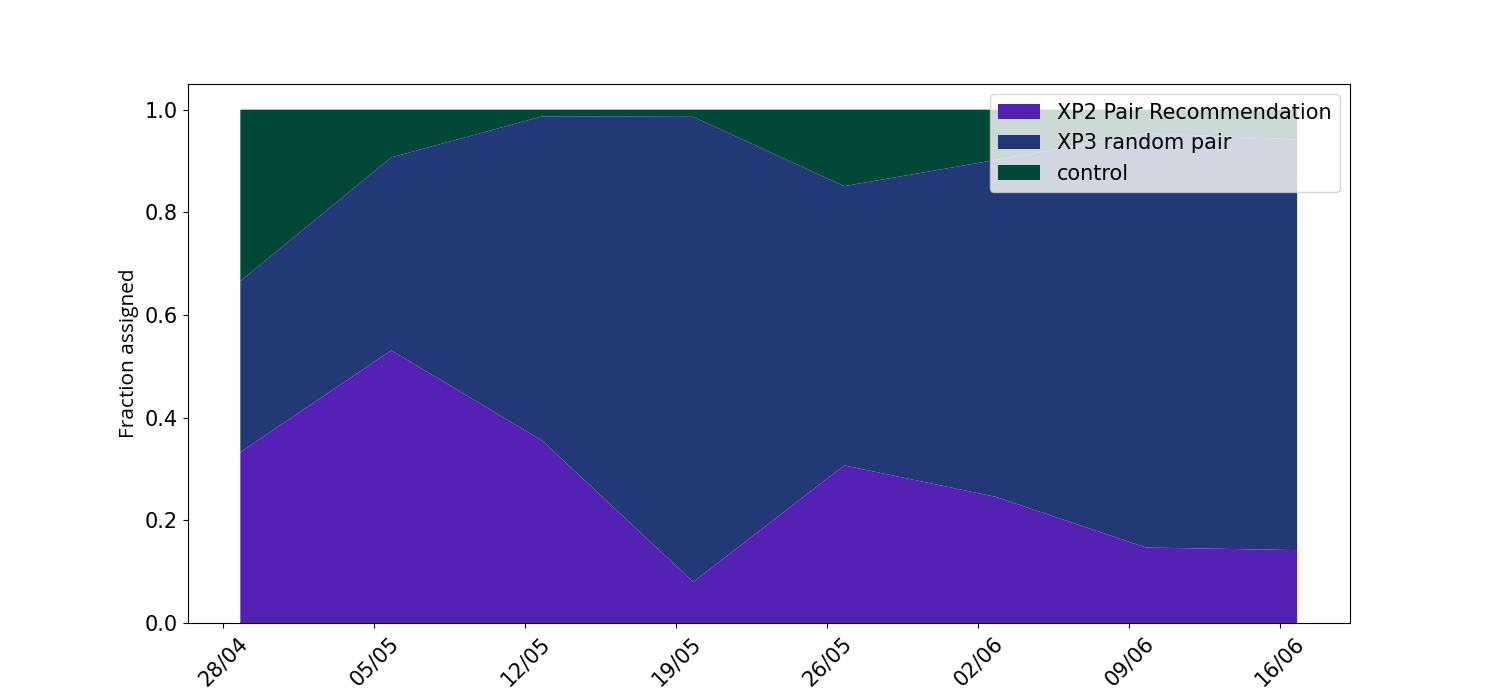}
  \caption{Fraction of participants assigned to each of XP3's adaptive intervention arms at each decision point.}
  \label{fig:xp3-split}
  \Description{Arm proportions plot for XP3.}
\end{figure*}

Figure \ref{fig:xp3-rewards} shows the evolution of the average reward per arm and decision point for XP3's adaptive intervention's arms, while Figure \ref{fig:xp3-split} the resulting proportions of assigned participants to each of the arms.

Table \ref{tab:xp3-sensitivity} contains the sensitivity classification of all arms to each of the contextual traits for XP3 (there is a non trivial relation between them as there are three arms and the sensitivities are obtained by normalizing between -1 and +1 the soft-thresholded values).

\begin{table*}[htpb]
  \caption{Sensitivity to context XP3.  Sensitivity categories are shown as follows: {\LARGE -} large negative, - medium negative, {\tiny -} small negative, empty negligible, {\tiny +} small positive, + medium positive, {\LARGE +} large positive. }
  \label{tab:xp3-sensitivity}
  \begin{tabular}{l c c c}
    \toprule
    &Personalized & Random & Control\\
      \toprule
    Expenditure previous 90 days &   {\tiny -} & {\tiny -}  & {\tiny -}\\
    Days between logins last 60 days & & \LARGE{-} & {\tiny -}\\
    In-app time last 30 days & {\tiny -} & {\tiny -} & {\tiny -}\\
    Days since last nudge &  {\tiny -} & {\tiny -} & -\\
    Days since first login & {\tiny -} & & - \\
    Days with order previous 30 days & \LARGE{-} & \LARGE{+} & \LARGE{-}\\
    Opened nudges last 14 days & {\tiny -} & {\tiny -} & {\tiny -}\\
    \midrule

\end{tabular}
\end{table*}

\begin{figure*}[htpb]
  \centering
  \includegraphics[width=0.45\linewidth]{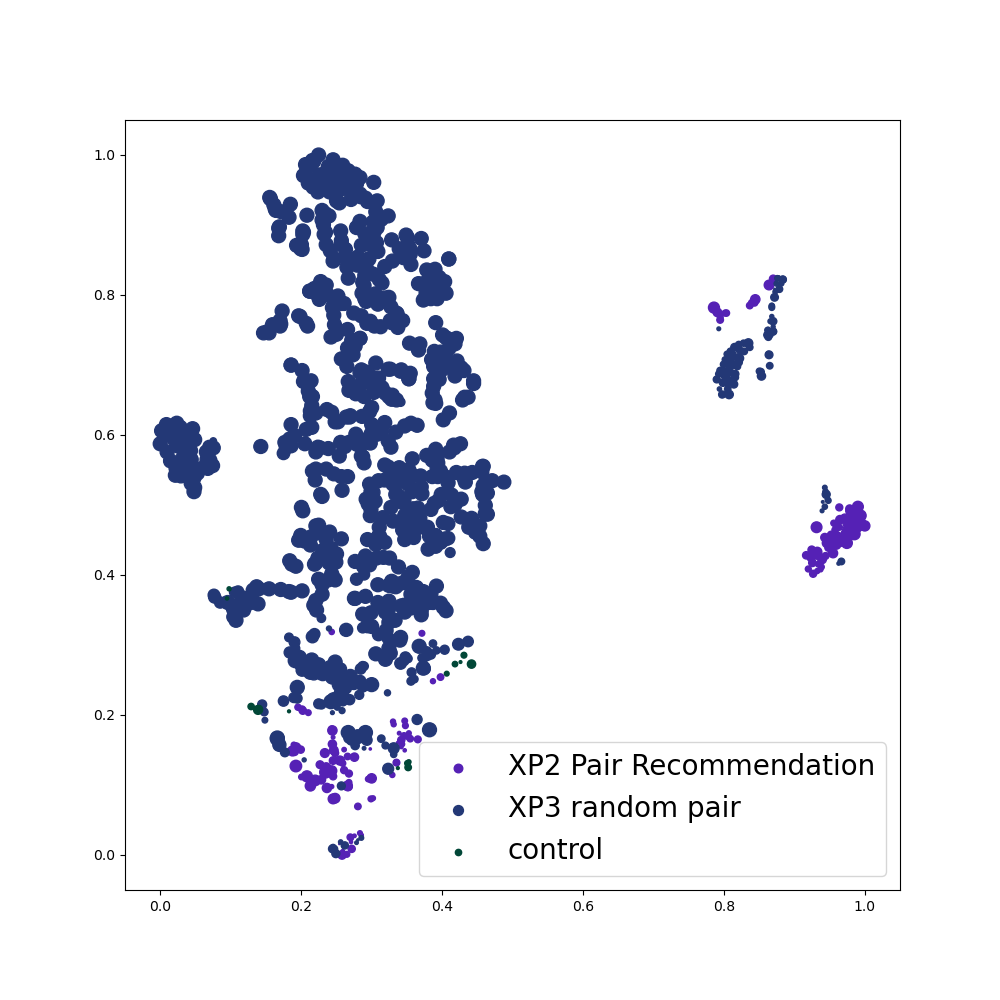}
  \includegraphics[width=0.45\linewidth]{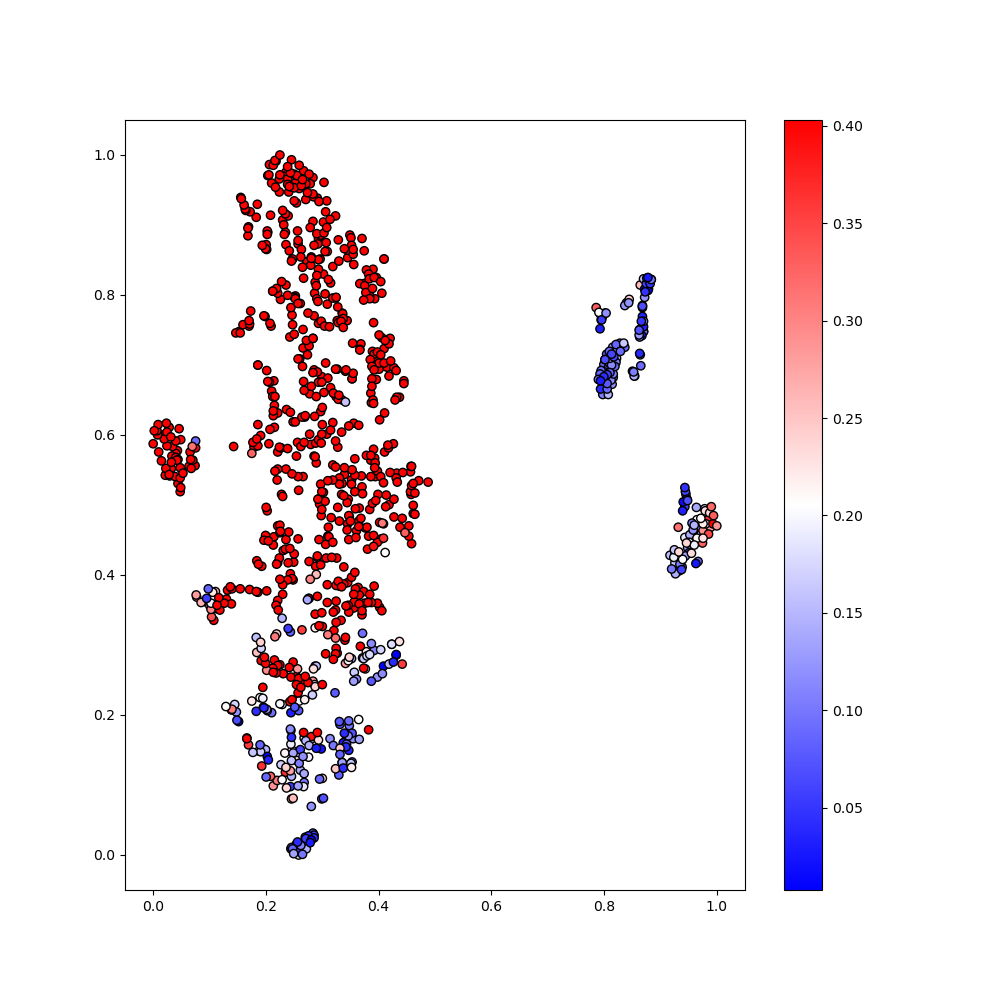}
  \caption{T-sne visualization in the contextual trait space for XP3. The figure to the left represents the best arm for each participant at the last decision point, with the size of the point proportional to its confidence (difference between the probabilities of the best arm and the second best arm), which is also plotted in the figure to the right. }
  \Description{T-sne plots for all users in XP3.}
  \label{fig:xp3-tsne}
\end{figure*}

Figure \ref{fig:xp3-tsne} shows the t-sne visualization in the contextual trait space of the best arm (left) and of its confidence computed as the difference in the probabilities of picking the best arm and second best arm (right) for XP3.

\subsection{LMM estimation}

Table \ref{tab:xp3-lmm-full} contains the estimated parameters, standard errors, and p-value of the term's Wald test for the full LMM described in Section \ref{sec:lmms} for XP3, with the addition that the two types of intervention (adaptive and non-adaptive) and of nudges (personalized or random item-pair) are considered explicitly. Additionally, a parameter to account for the lack of novelty the recommendations represented for those users that had already taken part in XP1 was also introduced. Table \ref{tab:xp3-lmm} is the equivalent table for the fitted LMM where all coefficients are significant.

\begin{table*}[htpb]
  \caption{Estimated parameters, standard error, and Wald's test p-value for the LMM with all terms for XP3 fitted to the weekly values.}
  \label{tab:xp3-lmm-full}
  \begin{tabular}{r c c c }
    \toprule
    &Coef. & Std.Err. & p-value\\
      \toprule
    Intercept & 40.730 & 4.575 & 0.000\\
    Adaptive intervention & -6.588 & 4.906  & 0.179\\
    Non-adaptive intervention & 0.194 & 8.708 & 0.982\\
    Nudged that week (personalized)  & 15.545 & 3.732 & 0.000 \\
    Nudged that week (random)  & 17.142 & 3.323 & 0.000\\
    XP1 participant & -10.640 & 4.014 & 0.008 \\
    Baseline expenditure & 901.561 & 15.106 & 0.000\\
    Week number & -4.759 & 0.602 & 0.000 \\
    Week number in non adaptive & 1.703 & 1.544 & 0.270 \\
    Week number in adaptive & -0.755 & 0.772 & 0.328\\
    \midrule

\end{tabular}
\end{table*}

\begin{table*}[htpb]
  \caption{Estimated parameters, standard error, and Wald's test p-value for the LMM with only significant terms for XP3 fitted to the weekly values.}
  \label{tab:xp3-lmm}
  \begin{tabular}{r c c c }
    \toprule
    &Coef. & Std.Err. & p-value\\
      \toprule
    Intercept &  38.643 & 3.814 & 0.000 \\
    Nudged that week (personalized)  & 11.307 & 3.212 & 0.000 \\
    Nudged that week (random)  & 11.723 & 2.613 & 0.000 \\
    XP1 participant & -10.506 & 4.029 & 0.009 \\
    Baseline expenditure & 905.241 & 15.093 & 0.000 \\
    Week number & -5.033 &  0.364 & 0.000 \\
    \midrule

\end{tabular}
\end{table*}

\subsection{Recommendation Success Analysis}

\begin{figure*}[htpb]
  \centering
  \includegraphics[width=0.45\linewidth]{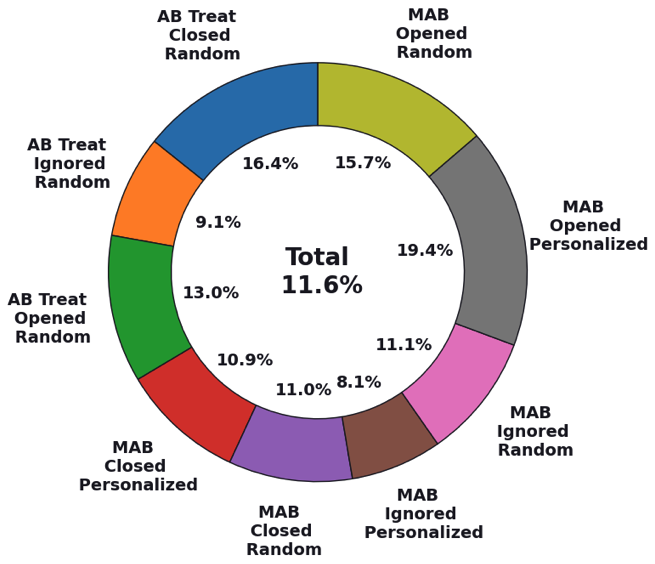}
  \includegraphics[width=0.45\linewidth]{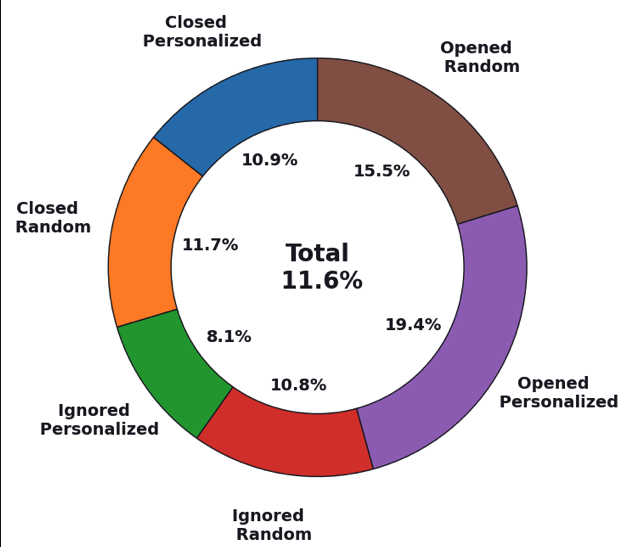}
  \caption{Breakdown by interaction (opened, closed or ignored), experimental group (MAB for adaptive intervention, AB for non-adaptive one), and recommendation type (personalized or random) of messages that lead to purchase in XP3 (11.6\% of all messages) to the left and by interaction and type of recommendation only (both experimental groups combined) to the right.}
  \Description{Recommendation success related pie charts for XP3. }
  \label{fig:xp3-success-all}
\end{figure*}

For XP3, for 11.6\% of the messages, considering both the adaptive and non-adaptive arms and both personalized and random recommendations, the item that had previously been infrequently purchased was ordered at a later time during the experiment. Figure \ref{fig:xp3-success-all} reflects the breakdowns by type of interaction, experimental group and recommendation type to the left and by interaction and recommendation type only to the right. 

\begin{figure*}[htpb]
  \centering
  \includegraphics[width=0.45\linewidth]{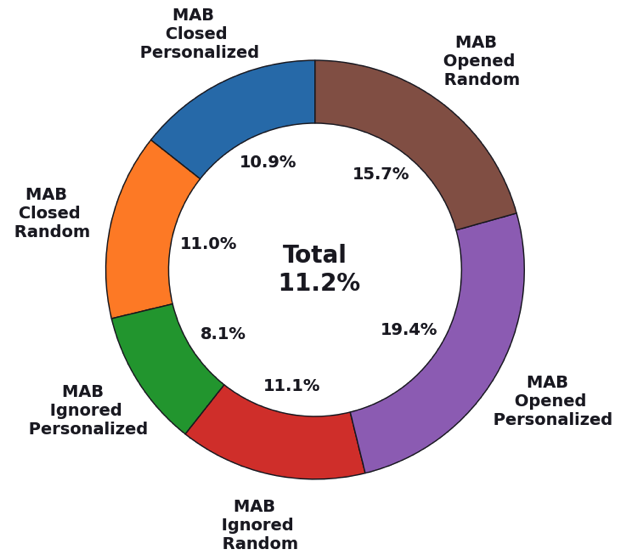}
  \includegraphics[width=0.45\linewidth]{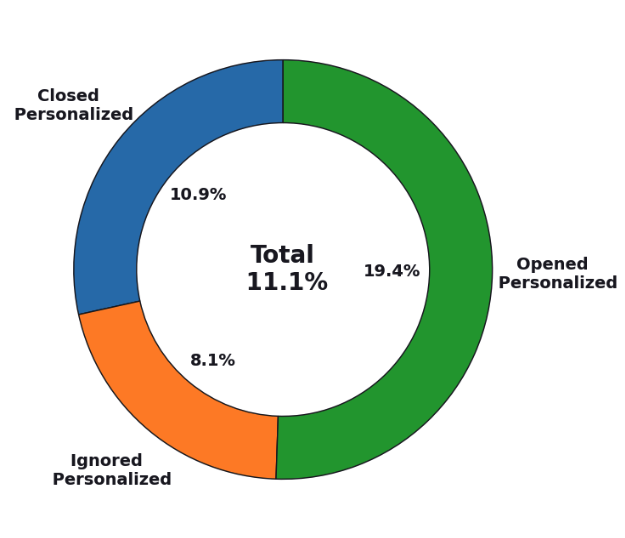}
  \caption{Breakdown by interaction (opened, closed or ignored), and recommendation type (personalized or random) of messages that lead to purchase in XP3's adaptive intervention group (11.2\% of all messages) to the left and for personalized recommendations only (11.1\% of the messages) to the right.}
  \Description{Recommendation success related pie charts for XP3.}
  \label{fig:xp3-success-adaptive}
\end{figure*}

\begin{figure*}[htpb]
  \centering
  \includegraphics[width=0.45\linewidth]{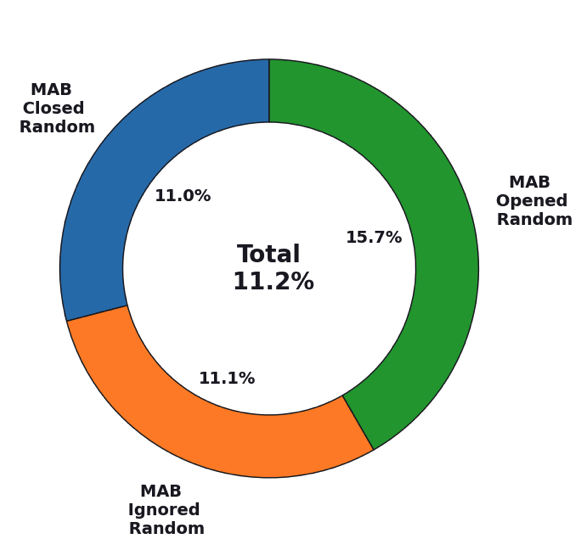}
  \includegraphics[width=0.45\linewidth]{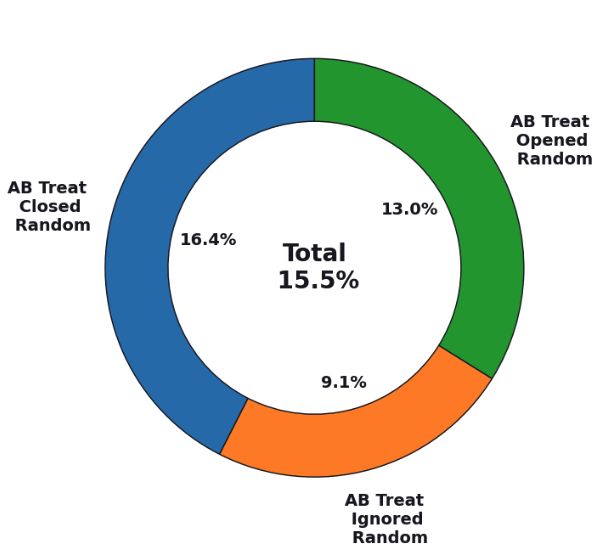}
  \caption{Breakdown by interaction (opened, closed or ignored) of random recommendations that lead to purchase in XP3's adaptive group (11.8\% of all messages) to the left and in XP3's non-adaptive group (15.5\%) to the right.}
  \Description{Recommendation success related pie charts for XP3.}
  \label{fig:xp3-success-random}
\end{figure*}

When focusing on the adaptive intervention experimental group only, 11.6\% of all messages ended in purchase, 11.1\% of the personalized ones and 11.2\% of the random ones. Figure \ref{fig:xp3-success-adaptive} shows the breakdown by interaction and recommendation type to the left and for personalized recommendations only by interaction to the left, while Figure \ref{fig:xp3-success-random}'s left plot for random recommendations.

\begin{figure*}[htpb]
  \centering
  \includegraphics[width=0.45\linewidth]{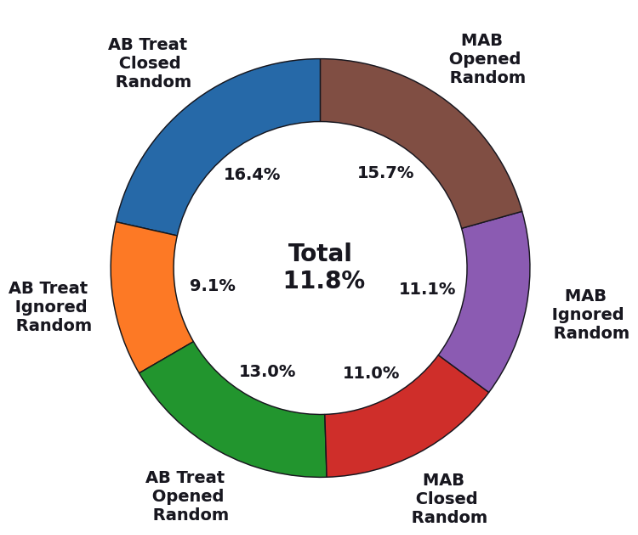}
  \includegraphics[width=0.45\linewidth]{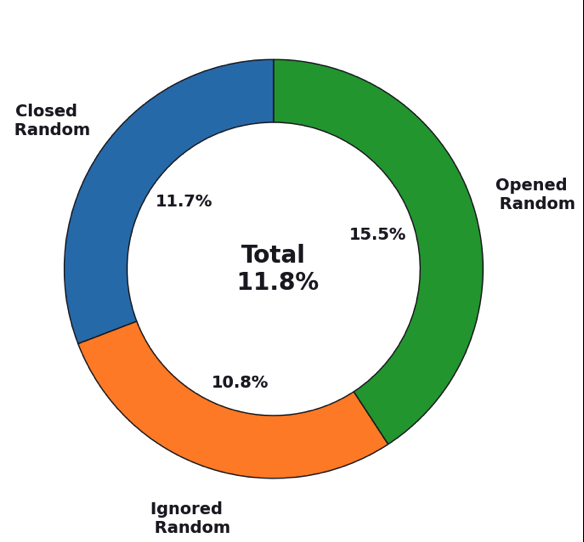}
  \caption{Breakdown by interaction (opened, closed or ignored), and experimental group (MAB for adaptive intervention, AB for non-adaptive one) of random recommendations that lead to purchase in XP3 (11.8\% of all messages) to the left and by interaction only to the right, with both the adaptive and non adaptive groups considered. }
  \Description{Recommendation success related pie charts for XP3.}
  \label{fig:xp3-success-random-all}
\end{figure*}

Figures \ref{fig:xp3-success-random-all} and Figure \ref{fig:xp3-success-random} explore in some detail the breakdowns for random recommendations exclusively. The former shows different breakdowns combining both the adaptive and non-adaptive experimental arms, while the latter shows the adaptive group to the left and the non-adaptive to the right.

\end{document}